\documentclass[10pt,twocolumn,letterpaper]{article}

\usepackage{wacv}

\usepackage[accsupp]{axessibility}
\usepackage{graphicx}
\usepackage{amsmath}
\usepackage{amssymb}
\usepackage{booktabs}
\usepackage{xcolor}
\usepackage{multirow}
\usepackage{adjustbox}
\usepackage{makecell}
\usepackage{float}
\usepackage{comment}

\usepackage[pagebackref,breaklinks,colorlinks]{hyperref}

\usepackage[capitalize]{cleveref}
\crefname{section}{Sec.}{Secs.}
\Crefname{section}{Section}{Sections}
\Crefname{table}{Table}{Tables}
\crefname{table}{Tab.}{Tabs.}

\def\wacvPaperID{1583} 
\def\confName{WACV}
\def\confYear{2024}

\begin{document}

\title{Spatially-Adaptive Hash Encodings For Neural Surface Reconstruction }

\author{Thomas Walker \and
Octave Mariotti \and \\ \hspace{-6cm}University Of Edinburgh\hspace{-6cm} \and Amir Vaxman \and Hakan Bilen 
    }


\maketitle

\begin{abstract}
   Positional encodings are a common component of neural scene reconstruction methods, and provide a way to bias the learning of neural fields towards coarser or finer representations. Current neural surface reconstruction methods use a “one-size-fits-all” approach to encoding, choosing a fixed set of encoding functions, and therefore bias, across all scenes. Current state-of-the-art surface reconstruction approaches leverage grid-based multi-resolution hash encoding in order to recover high-detail geometry. We propose a learned approach which allows the network to choose its encoding basis as a function of space, by masking the contribution of features stored at separate grid resolutions. The resulting spatially adaptive approach allows the network to fit a wider range of frequencies without introducing noise. We test our approach on standard benchmark surface reconstruction datasets and achieve state-of-the-art performance on two benchmark datasets.
\end{abstract}

\section{Introduction}
\label{sec:intro}
Surface reconstruction is a long standing problem in the fields of computer vision and graphics. Traditional methods, such as multi-view stereo (MVS) and structure-from-motion (SfM) \cite{schoenberger2016sfm, furukawa2009accurate, snavely2006photo}, have been the standard for reconstructing 3D surfaces from 2D images. 
These approaches rely heavily on precise feature matching and depth estimation to extract sparse representations which are prone to fail on regions of uniform texture or non-Lambertian material. Errors accumulated in the feature matching and subsequent surface reconstruction often lead to noisy surfaces and missing regions. \footnotetext{OM and HB were supported by the EPSRC Visual AI grant EP/T028572/1}

Neural scene reconstruction has made significant strides since the introduction of Neural Radiance Fields (NeRF) \cite{mildenhall2021nerf}, which pioneered the use of neural networks to generate 3D scenes from 2D images. NeRF models learn to predict volumetric scene representations by encoding the geometry and appearance of complex scenes using light fields, leading to high-quality 3D scene reconstructions. Despite NeRF's success, it is poorly suited to accurately reconstruct surfaces because it represents scenes as continuous volumetric densities rather than exact geometric surfaces. This limitation can lead to blurred or imprecise boundaries when modeling objects with sharp edges or fine textures, reducing its effectiveness for tasks requiring precise surface reconstructions. Subsequent work attempted to address this limitation by representing scenes as the level set of a Signed Distance Function (SDF), which allows for precise surface modeling~\cite{wang2023neus, li2023neuralangelo, yariv2021volume}. These methods use the SDF to directly encode the geometry of a scene, capturing surfaces more accurately by treating the scene's surface as the zero-crossing of the volumetric function. Notably, NeuS~\cite{wang2023neus} demonstrated that optimization of a neural signed distance function can be achieved using NeRF's popularized volume rendering framework~\cite{mildenhall2021nerf}. By learning surfaces directly from images, these methods avoid the error accumulation seen in traditional approaches and have resulted in new state-of-the-art surface reconstruction algorithms \cite{li2023neuralangelo, wang2023neus}.

\begin{figure}[t]
    \centering
    \includegraphics[width=1\linewidth]{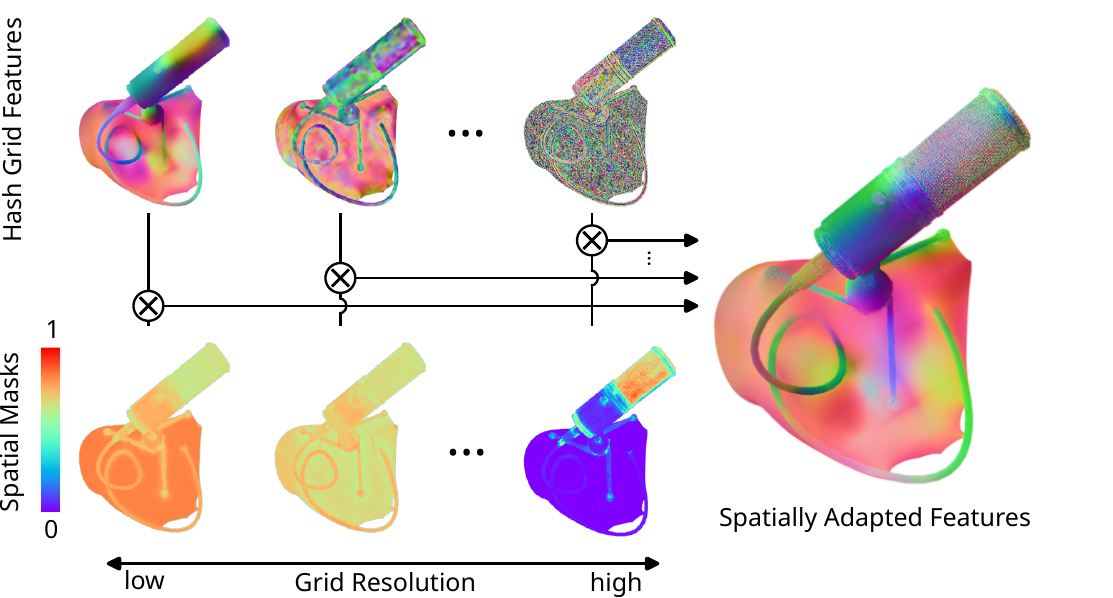}

    \caption{In order to modulate typical hash grid features fields (top row), we propose to jointly learn a scalar field for each grid resolution (bottom row). Combining them yields a spatially-adaptive encoding which allows fine features in high-detail regions (microphone tip) while preserving lower frequency embeddings on smoother regions (microphone body).}
    \label{fig:splash}
\end{figure}

A core component of all neural reconstruction methods is the use positional encodings to facilitate learning fine details, since neural networks are biased towards smooth functions on low dimensional domains~\cite{tancik2020fourier}. NeRF popularized the use of sinusoidal positional encodings, however, this approach still fails to capture the finest surface details. 

While increasing the frequency of the encoding in attempt to learn both low and high-frequency components simultaneously is possible in theory, such a strategy leads to unstable optimization and unwanted noise~\cite{wang2022hf, hertz2021sape}. SAPE~\cite{hertz2021sape} introduced spatially-adaptive sinusoidal encodings, enabling neural fields to dynamically select their positional encoding frequencies. This allows the model to effectively capture scenes with both fine and coarse details, providing more stable and flexible fitting across varying feature scales. 

An alternative to sinusoidal encodings has been proposed as feature grid-based encodings \cite{li2023neuralangelo, takikawa2021nglod, liu2020neural}. These methods discretize the scene into a grid, storing features at grid points that can be interpolated to represent the scene more efficiently. Recently, hashing functions have been employed to improve the scalability of grid-based encoding approaches, reducing memory usage while maintaining high resolution, ultimately achieving impressive surface accuracy~\cite{li2023neuralangelo}. 
Nonetheless, the resolution of grids used by these works is fixed and global (\ie does not depend on the spatial location).

In this work, we propose a spatially-adaptive formulation of multi-resolution hash grids, which allows the model to adaptively select the resolution of grid used based on the complexity of the scene. By dynamically adjusting the encoding bias at different spatial regions, our approach balances the need for high-detail surface reconstruction in intricate areas without introducing noise in smoother regions (see \Cref{fig:splash}). Our method contributes more precise and flexible reconstructions across a wide range of scene complexities, achieving state-of-the-art surface reconstructions.

\section{Related Work}
\label{sec:relwork}
\subsection{Neural Reconstruction}
Neural reconstruction has progressed substantially with the use of neural fields. NeRF popularized the use of volumetric neural functions for 3D scene rendering, but its reliance on density fields often leads to poor surface quality, making it unsuitable for tasks requiring exact, detailed geometry. Alternatively, explicitly representing geometry through neural parametric surfaces~\cite{walker2023explicit, groueix2018papier, williams2019deep, bednarik2020shape}, while more efficient, are either limited to spherical topology or introduce discontinuities at parametric patch boundaries. In contrast, recent techniques~\cite{yariv2021volume, takikawa2021nglod, li2023neuralangelo} define surfaces implicitly as the zero-level set of an SDF. By implicitly representing the surface, these approaches can attain surfaces of arbitrary topology with clearly defined surfaces. We build and improve upon these approaches to attain the most accurate implicit neural surfaces \cite{li2023neuralangelo}.

\subsection{Positional Encodings}
Positional encoding is a critical component in optimizing neural networks on low-dimensional domains. The transformation of input coordinates into a higher-dimensional space facilitates the neural network's ability to capture intricate details and spatial relationships \cite{tancik2020fourier}. Various types of positional encoding have been explored and employed in literature to enhance the performance of neural networks in tasks such as surface reconstruction \cite{gropp2020implicit,wang2023neus, wang2022hf}, image synthesis \cite{hertz2021sape, muller2022instant}, and natural language processing \cite{vaswani2017attention}.
\paragraph{Frequency-based Encodings}  

involve encoding the input coordinates using sinusoidal functions of varying frequencies. This method significantly improved the capability of neural networks to represent detailed and complex surfaces \cite{wang2023neus, wang2022hf}. Subsequent improvements such as random Fourier features leverage random Gaussian matrices to transform input coordinates \cite{tancik2020fourier}. This stochastic approach addressed artifacts produced by axis-aligned sinusoidal functions and enables a more flexible representation of spatial patterns. Frequency-based positional encodings have been generalized to other geometric domains, for example, spherical harmonics provide an intrinsic basis of periodic functions on the sphere, making them particularly useful for encoding directions and orientations. In the context of surface reconstruction, spherical harmonics are used to encode the angular information of surface normals or light directions, enhancing the neural network's ability to model complex lighting and shading effects \cite{li2023neuralangelo, verbin2022ref,mueller2022instant}.\\

\paragraph{Grid-Based Encodings} offer an alternative to frequency-based encodings through the use of feature grid based approaches, also known as voxel grids, involve dividing the space into (sometimes multiple) grids, where each grid vertex stores a trainable feature vector. The input coordinates are mapped to these feature vectors based on their spatial location within the grid, typically through a choice of interpolation function \cite{takikawa2021nglod, liu2020neural, mueller2022instant}. These methods delegate a large portion of learning task to the feature-grids themselves, allowing for the use of smaller and more efficient MLPs, and hence improving training and inference time. As a big drawback, grid-based approaches have significant scalability issues. For high-resolution reconstructions, the memory requirements can become prohibitive, as the number of voxels grows cubically with the resolution. Initial attempts to address memory usage include progressive grid pruning approaches based on octrees \cite{takikawa2021nglod} and sparse grids \cite{liu2020neural}. However, these approaches require complex training schemes which are difficult to implement efficiently on GPUs due to their dynamic data structures. To address these challenges, M{\"u}ller et. al.~\cite{mueller2022instant} leveraged a fixed-size hash table to store feature vectors, mapping input coordinates through a hash function. The fixed cache size simplifies memory management, significantly reducing memory usage and training times while maintaining high-resolution detail, thereby addressing the scalability issues faced by grid-based methods. 

\subsection{Spatially-Adaptive Encodings}
All choices of positional encoding impart a bias on the representation attained. Namely, the use of high-frequency encoding functions (or analogously, high-density grids) facilitates fitting to signals with fine detail, but sacrifices the smoothness prior of neural fields. Typically, this bias is tuned as a hyper-parameter, which can be both time consuming and ill-suited to complex scenes with both ``smooth'' and ``fuzzy'' properties. In many cases, attempts to capture fine detail with high-frequency encodings makes the optimization sensitive to noise, degrading performance and leading to stability issues \cite{wang2022hf}. To address this, spatially-adaptive encodings have been employed for target signals which are inconsistent across space. For example, Ref-NeRF \cite{verbin2022ref} and NRFF \cite{Han_2023_CVPR} use spatially-adaptive view direction encodings to handle scenes with regions of both specular and diffuse reflectance distributions. Most related to our approach, SAPE \cite{hertz2021sape} employ a spatially-adaptive encoding of spatial coordinates for image synthesis and 3D occupancy regression tasks. This approach enables fitting a wide range of frequencies without sacrificing training stability or introducing a high-frequency bias to smooth regions. However, their approach is still subject to the drawbacks of frequency-based encodings. We propose spatially-adaptive hash-grid encodings, combining the advantages of grid-based and spatially-adaptive methods. This combination is particularly attractive, since hash-compression artifacts occur most frequently in the finest grids, presenting an incentive to ignore their usage where possible.

\section{Methodology}
\label{sec:method}
We build our method on Neuralangelo \cite{li2023neuralangelo} and augment it with our spatially adaptive encoding formulation.
We first outline core components of Neuralangelo in ~\cref{prelim}, then in~\cref{ourmethod} we present our spatially-adaptive hash encoding methodology.  

\subsection{Preliminaries}
\label{prelim}
\textbf{Volume Rendering SDFs}
Neural surfaces can be represented as the level-set of a volumetric function which is parameterized by a neural network. Signed-distance-functions (SDFs) are one of the most common place surface representations, representing geometry by its zero-level set, i.e., $\mathbf{S} = \{ \mathbf{x} \in \mathbb{R}^3 \mid SDF(\textbf{x}) = 0 \}$, where $SDF(\mathbf{x})$ is the SDF value. It was demonstrated in NeuS \cite{wang2023neus} that a conversion between volume density predictions in NeRF and SDF representations using a logistic function allowed for optimization with volume rendering. The rendered color $\hat{\textbf{c}}$ of a pixel is approximated by the sum:
\begin{equation}
\hat{\mathbf{c}} = \sum_{i=1}^N w_i \mathbf{c}_i, \quad \text{where} \quad w_i = T_i \alpha_i \label{eqn:volume_rendering},
\end{equation}
where $c_i$ is the color of coordinate $x_i$, $\alpha_i$ is the opacity of the $i$-th ray segment, and $T_i = \prod_{j=1}^{i-1} (1 - \alpha_j)$ is the accumulated transmittance. For the opacity value $\alpha_i$, given a coordinate $\mathbf{x}_i$ and its SDF value $SDF(\mathbf{x}_i)$, we compute:
\begin{equation}
\alpha_i = \mathrm{max} \left( \frac{\Phi_s(SDF(\mathbf{x}_i)) - \Phi_s(SDF(\mathbf{x}_{i+1}))}{\Phi_s(SDF(\mathbf{x}_i))}, 0 \right), \label{eqn:sdf_opacity}
\end{equation}
where $\Phi_s$ is the sigmoid function. When provided with a specific camera position and ray direction, the volume rendering process integrates the color radiance of sampled points along the ray. The SDF, $\sigma_i$, and color, $c_i$, of each sampled point are predicted using a coordinate-based MLP.  

\textbf{Numerical Gradients} As illustrated in Neuralangelo \cite{li2023neuralangelo}, the use of numerical gradients can help enforce surface consistency. To compute surface normals, two additional SDF samples are taken along each coordinate axis with a step size $\epsilon$. For a sampled coordinate $\textbf{x}=(x, y, z)$, we compute the components of the derivative (here the $x$-component),
\begin{equation}
\nabla_x SDF(\textbf{x}) \approx \frac{SDF( \textbf{h}(\textbf{x} + \boldsymbol{\epsilon}_x)) - SDF(\textbf{h}(x - \boldsymbol{\epsilon}_x))}{2\epsilon},
\end{equation}
where $\boldsymbol{\epsilon}_x = [\epsilon,0,0]$. For a large enough step size $\epsilon$, surface normal computations use features from neighbouring hash-grid cells, and hence allow multiple cells to be optimized simultaneously. The effect of this can be considered a smoothing operation, as when used alongside the eikonal gradient regularizer, it will enforce surface normal consistency at a larger scale. This insight is leveraged to create a coarse-to-fine optimization by scheduling the $\epsilon$ parameter to approach zero (and hence the analytical analogue) across training.\\

\subsection{Spatially-Adaptive Hash Encoding}
\label{ourmethod}

 To implement spatially-adaptive features, we propose to split the standard feature hash grids in two separate components. The first is an implicit surface representation, namely a signed distance function (SDF), akin to that used in related approaches\cite{mueller2022instant, li2023neuralangelo}. The second, smaller, hash grid is used for a neural field representing a spatially varying mask. This mask is used to adapt the encoding received by the SDF. 
 
 Given an input coordinate $\textbf{x} = (x,y,z)$ we return interpolated feature vectors $\textbf{f}$ from the SDF multi-resolution hash grid and $\textbf{g}$ from the spatial mask hash grid. We then feed the feature $\textbf{g}$ into a shallow multi-layer perceptron (MLP) with sigmoid activation, producing a spatial mask $\textbf{s}$. The mask is a vector of $N$ scalar values $s_l(\textbf{x}) \in (0,1)$ for each level $l \in [1, N]$ of the SDF hash-grid. This serves to weight the contribution of the different resolutions of the hash-encoding which the SDF MLP receives. If we consider  $\textbf{f}$ as the concatenation of features from grid levels $\textbf{f} = [\textbf{f}_1, \dots,  \textbf{f}_N$], then we compute the final hash encoding as,

\begin{equation}
    \textbf{h}(\textbf{x}) = [s_1(\textbf{x}) \cdot \textbf{f}_1, \dots, s_N(\textbf{x}) \cdot \textbf{f}_N].   
\end{equation}
Since this mask is a function of space, the network can silence the contribution of any resolution hash grid in a spatially adaptive way. For example, removing the high-frequency bias induced by the finest hash grids in smooth regions of the scene. 

\textbf{Progressive Encoding} 
In analogy with SAPE \cite{hertz2021sape}, we employ progressive encoding scheme which introduces the high-frequency encodings over time. More precisely, alongside scheduling numerical gradients, we progressively unveil finer grids to the SDF network throughout optimization. The reasons for this are two-fold. Firstly, if all resolutions were active upon initialization, then fine hash grids would be smoothed by the use of large $\epsilon$, and will need to ``relearn'' details with small $\epsilon$ in the latter stages of training. As noted in Neuralangelo \cite{li2023neuralangelo}, this can prevent fine details from ever being learned. Secondly, if all grid resolutions are active from the start of training, then the spatial mask will learn to mask out the finer hash grids which aren't required early in training. If this behaviour converges, then the spatial mask network will never be used in the latter stages of optimization for attaining fine details. In practice, we prevent this by blocking gradients to the spatial masks of inactive grid resolutions.

\subsection{Optimization}
The network is overfitted by using a pixel-wise L1 loss on the input images, $\mathbf{I}$, and the rendered images, $\hat{\mathbf{I}}$:
\begin{equation}
L_{\mathbf{rgb}} = \| \hat{\mathbf{I}} - \mathbf{I} \|_1. \label{eqn:color_loss}
\end{equation}

In addition to this we use the geometric regularization first proposed by Gropp et al. \cite{gropp2020implicit}. Since a signed distance function should change linearly along the direction perpendicular to the level sets we encourage this property by regularizing the surface with an eikonal loss,
\begin{equation}
    L_{\text{eik}} = \frac{1}{N} \sum_{i=1}^N \left( \|\nabla SDF (\mathbf{x}_i) \|_2 - 1 \right)^2 \label{eqn:eikonal_loss}.
\end{equation}

Similarly to Neuralangelo \cite{li2023neuralangelo}, we additionally encourage smoothness by regularizing the mean curvature of the Signed Distance Function (SDF). This mean curvature is calculated using the discrete Laplacian, similar to how surface normals are computed. The curvature loss, $L_{\text{curv}}$, is defined as:

\begin{equation}
L_{\text{curv}} = \frac{1}{N} \sum_{i=1}^N \left | \nabla^2 SDF (\mathbf{x}_i) \right |. 
\end{equation}
The total loss is expressed as a weighted sum of the losses:
\begin{equation}
 \mathcal{L} = L_{\text{rgb}} + w_{\text{eik}} L_{\text{eik}} + w_{\text{curv}} L_{\text{curv}} \,.   
\end{equation}
Both spatial mask and SDF networks are trained jointly in an end-to-end manner. For our experiments we use values of $0.1$ and $5\times 10^{-4}$ for $w_{\text{eik}}$ and $ w_{\text{curv}}$ respectively.     
\begin{figure*}[h!]
    \centering
    \begin{tabular}{ccccc}
        & Input & Ours & Neuralangelo \\ 
        
        \rotatebox{90}{\parbox[t]{3cm}{\hspace*{\fill}Scan 37 \hspace*{\fill}}}\hspace*{5pt} & 
        \includegraphics[trim=5cm 5cm 5cm 5cm, clip, width=0.30\linewidth]{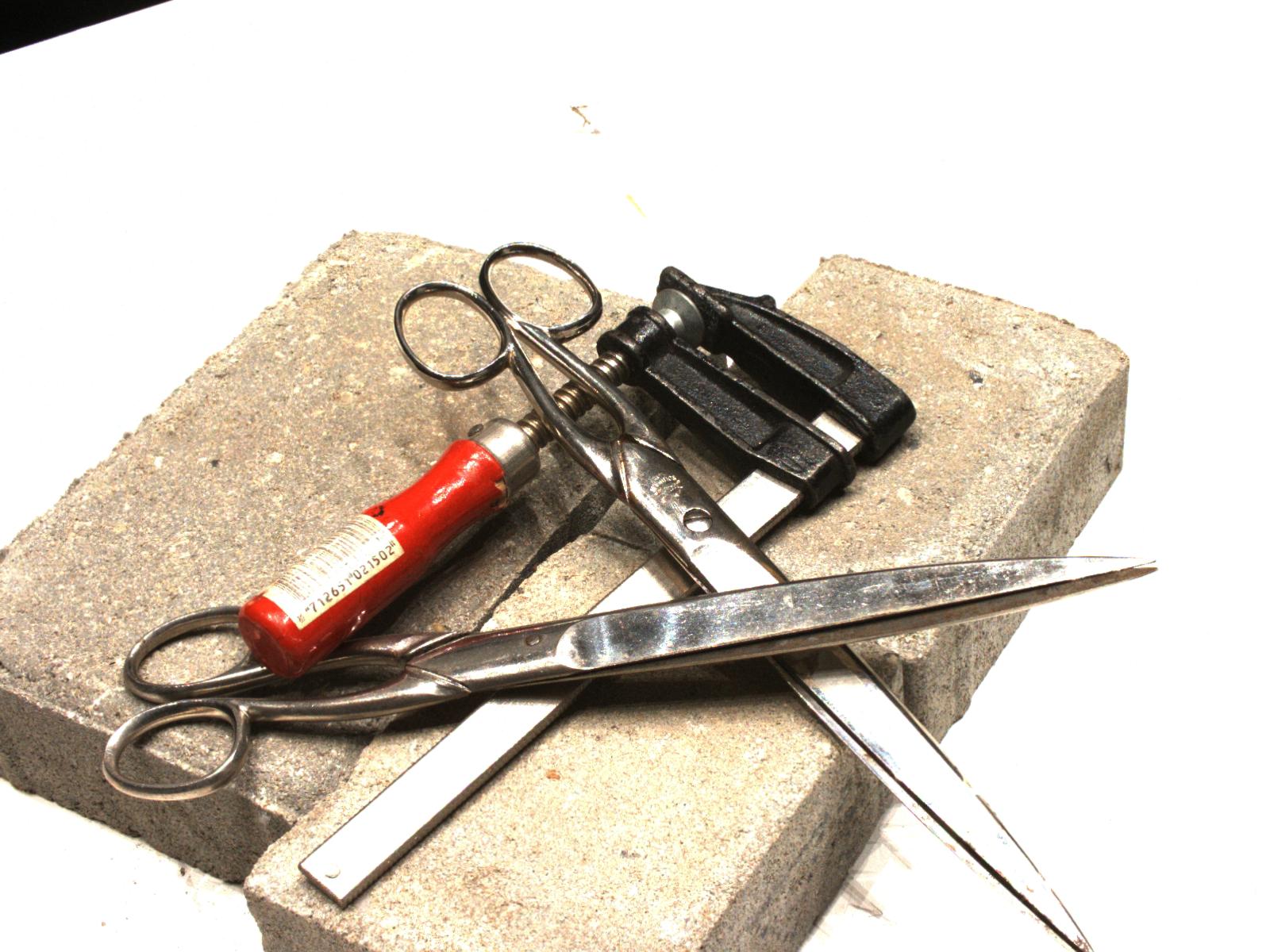} & 
        \includegraphics[trim=10cm 10cm 10cm 10cm, clip, width=0.30\linewidth]{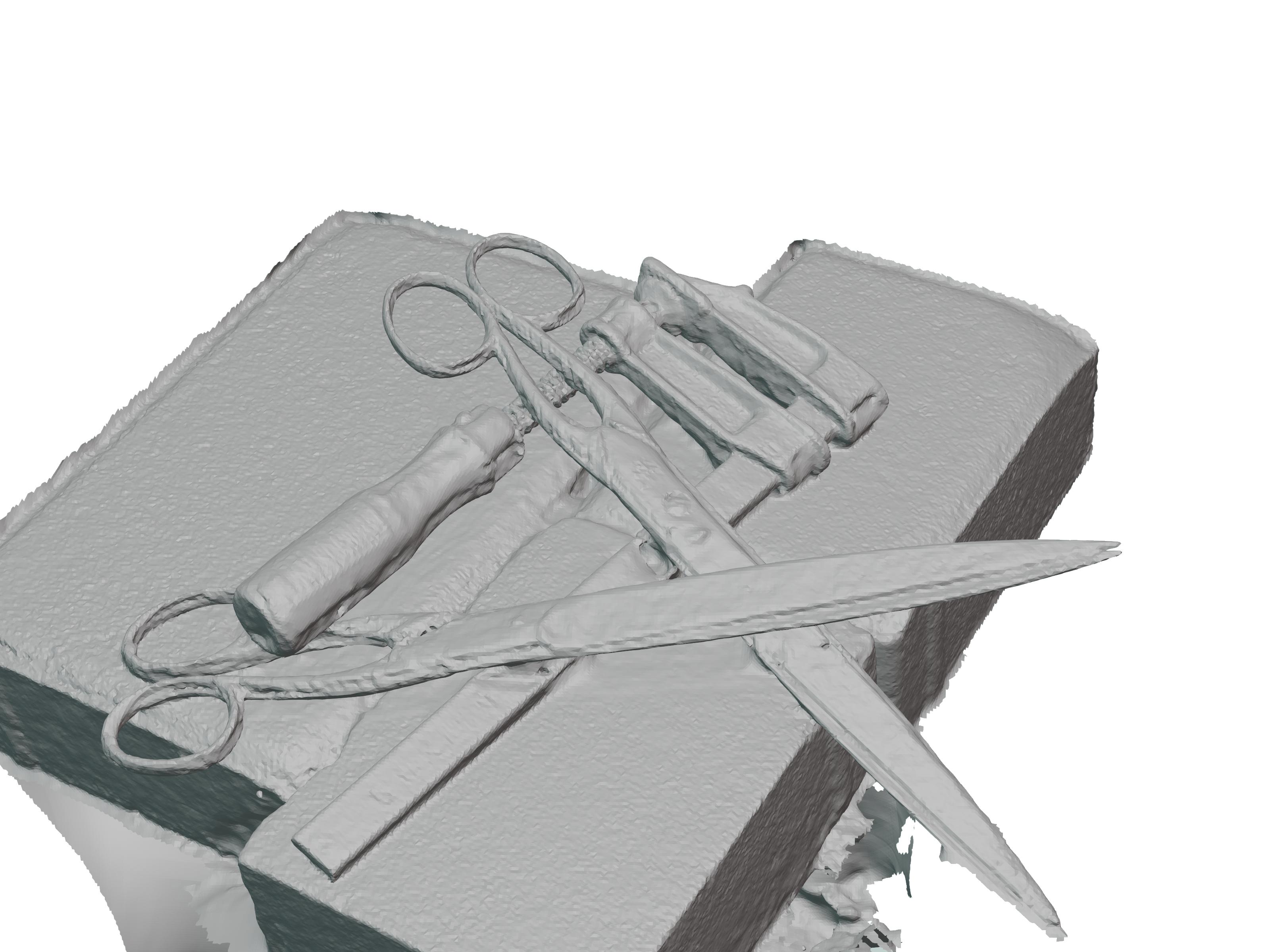} & 
        \includegraphics[trim=10cm 10cm 10cm 10cm, clip, width=0.30\linewidth]{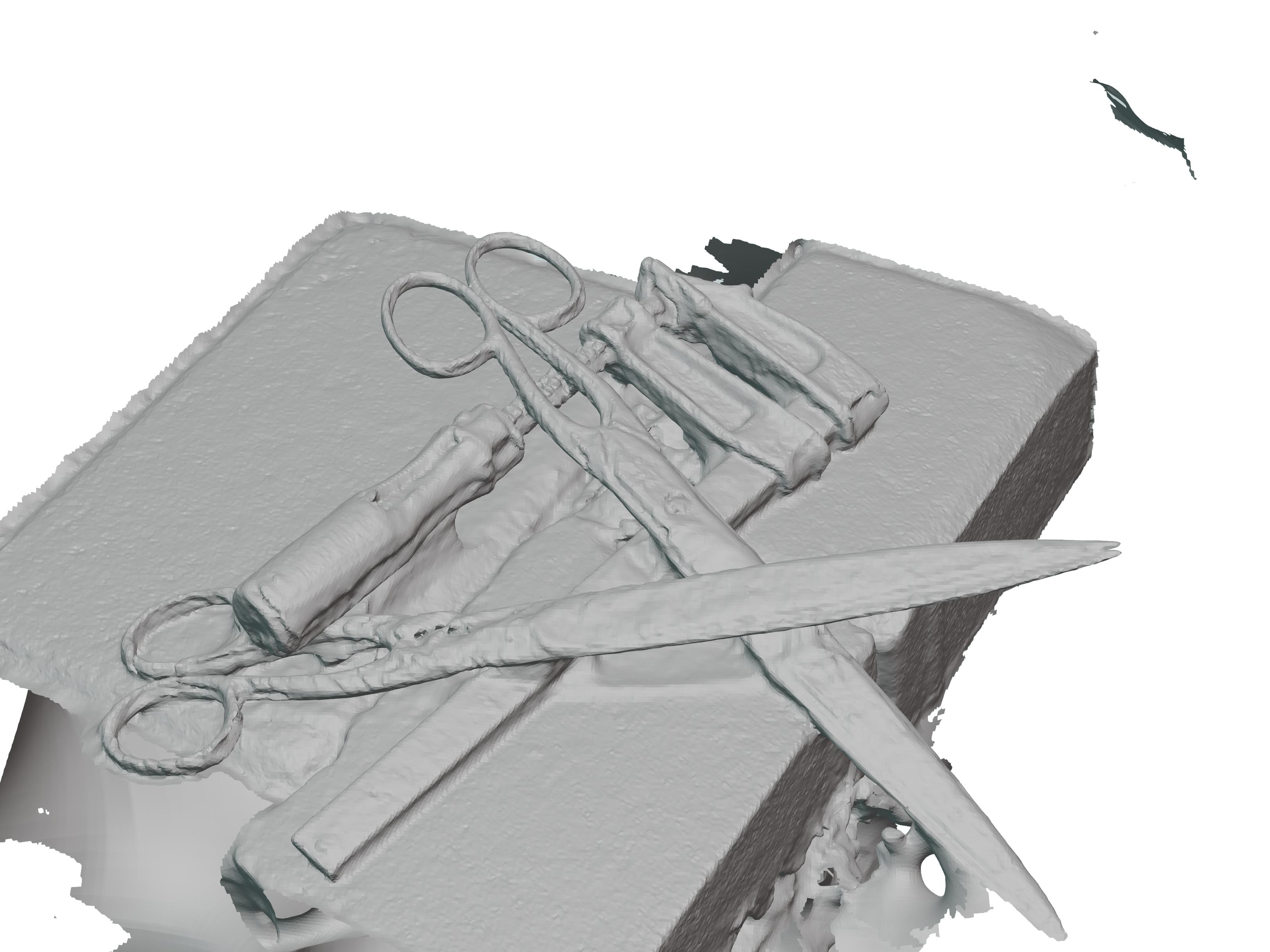}\\
        
        \rotatebox{90}{\parbox[t]{3cm}{\hspace*{\fill} Scan 97\hspace*{\fill}}}\hspace*{5pt}& 
        \includegraphics[width=0.3\linewidth]{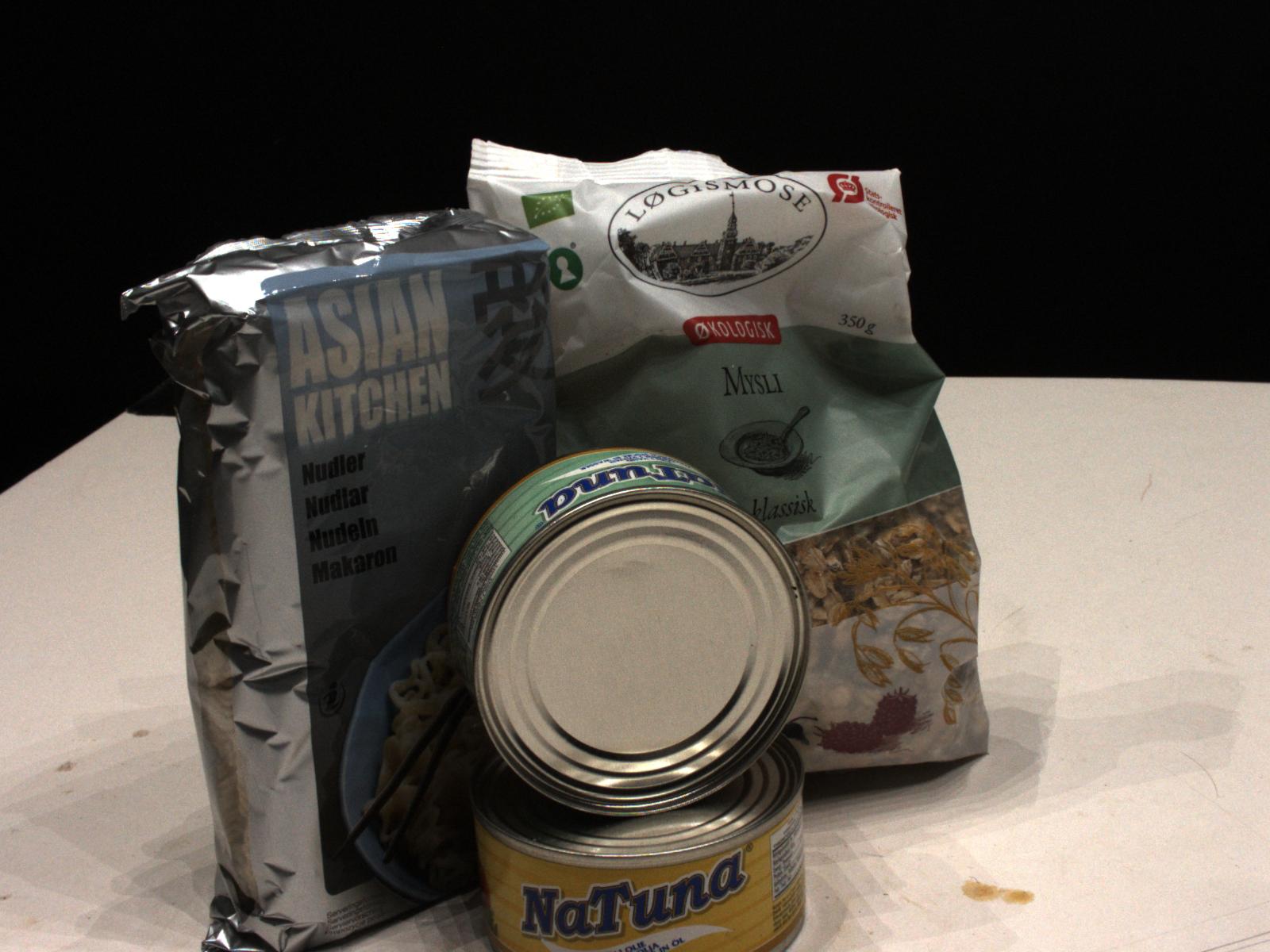} & 
        \includegraphics[width=0.3\linewidth]{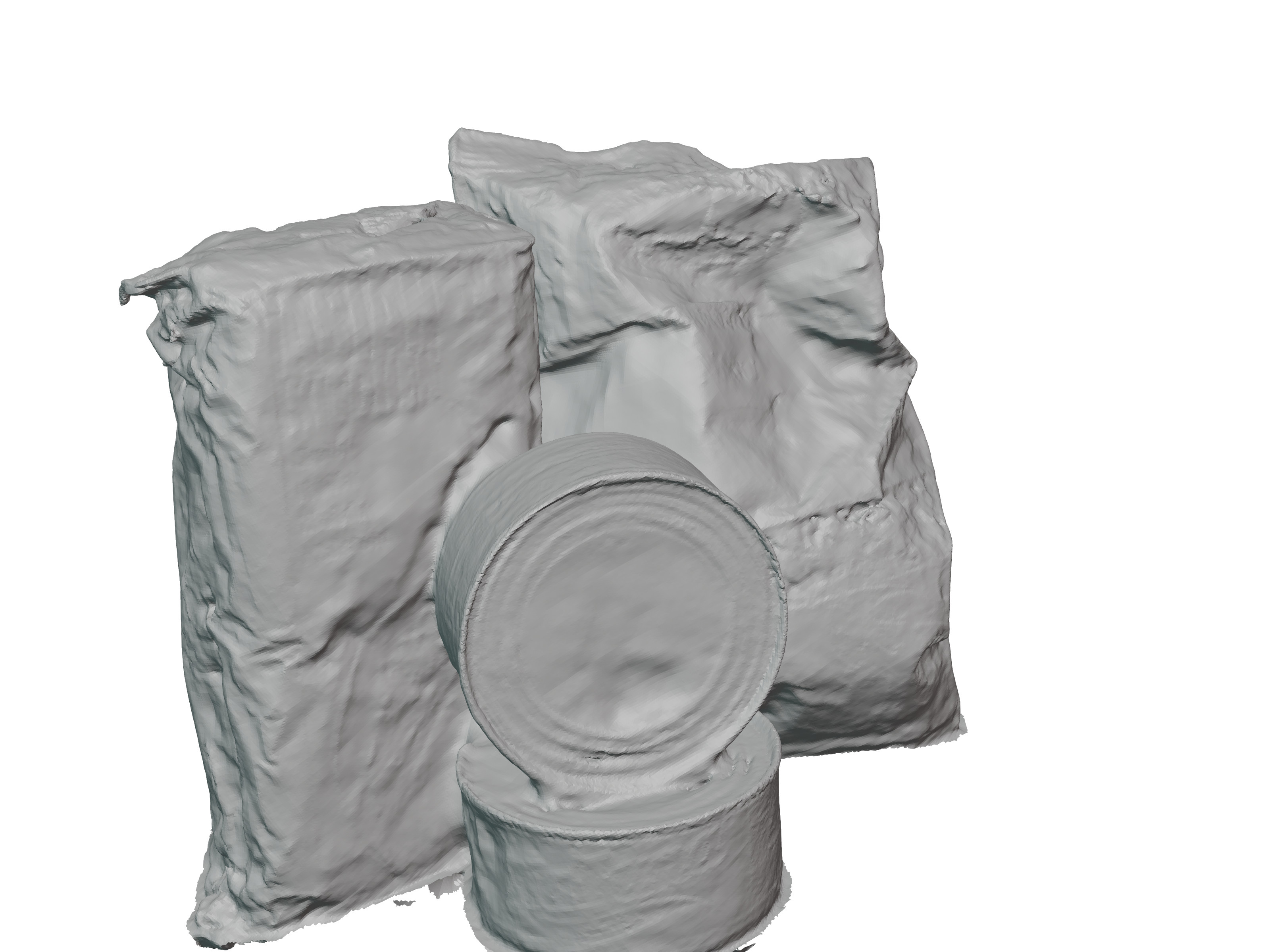} & 
        \includegraphics[width=0.3\linewidth]{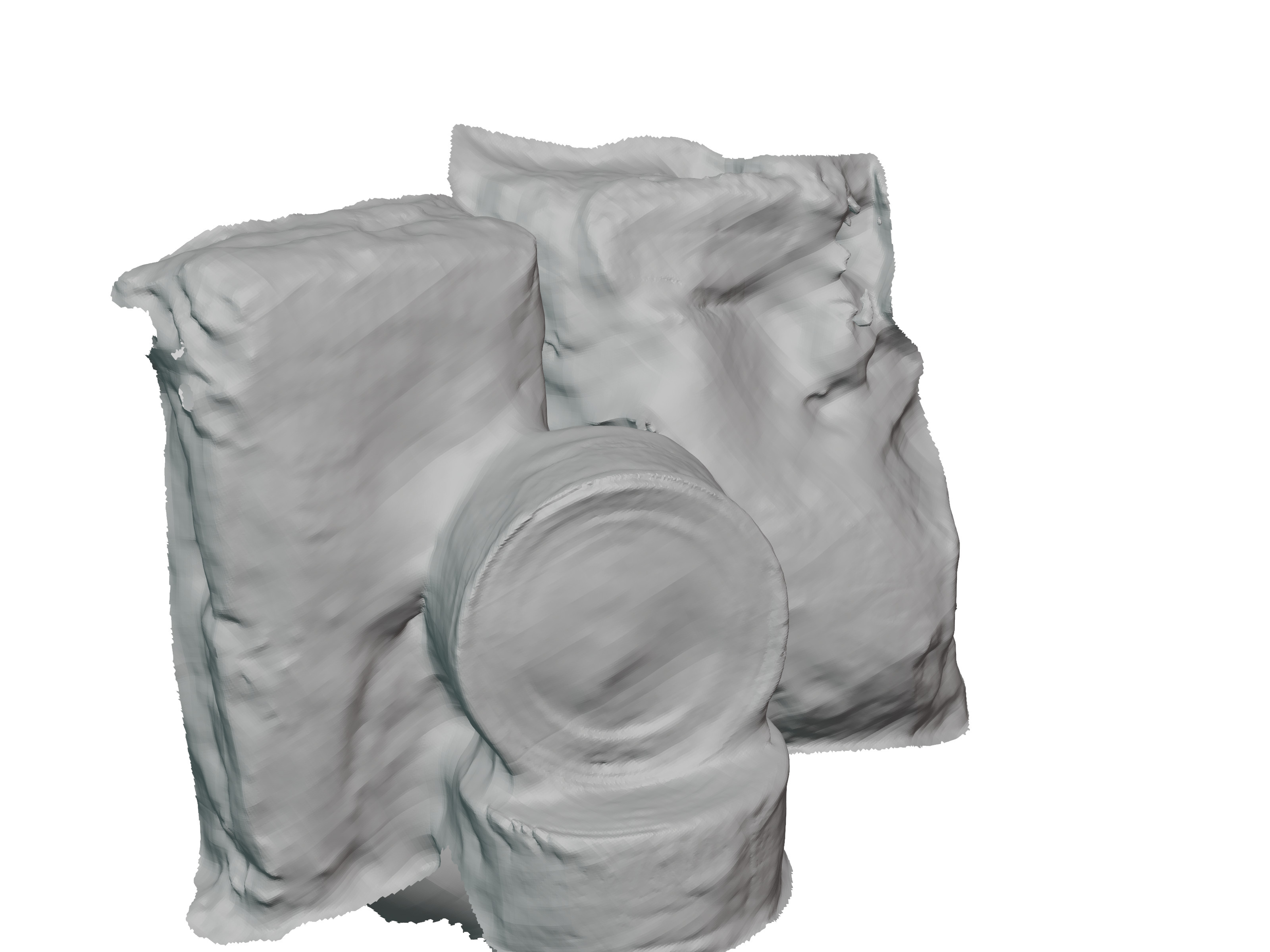} \\

        \rotatebox{90}{\parbox[t]{3cm}{\hspace*{\fill}Scan 65\hspace*{\fill}}}\hspace*{5pt} & 
        \includegraphics[width=0.3\linewidth]{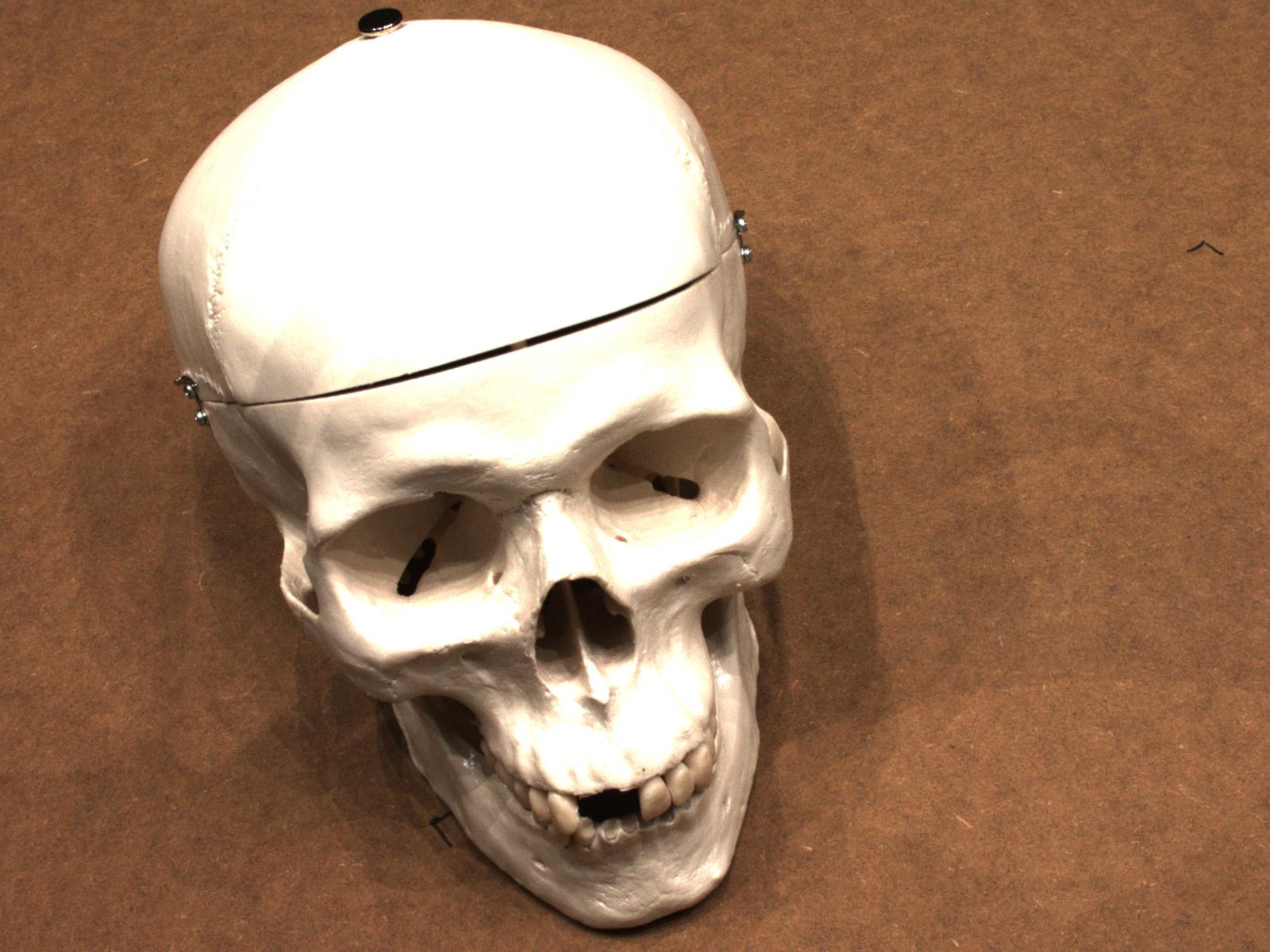} & 
        \includegraphics[width=0.3\linewidth]{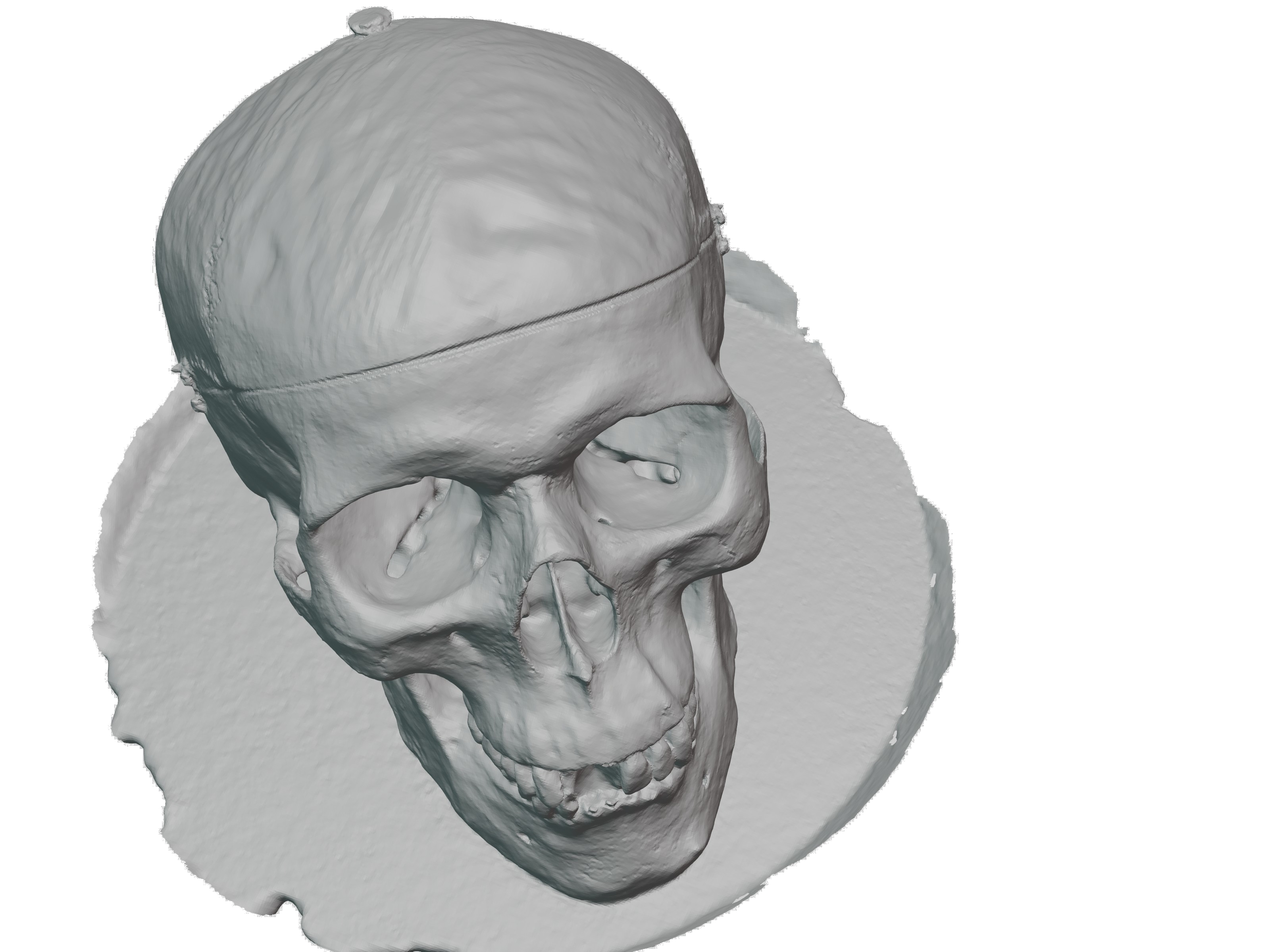} & 
        \includegraphics[width=0.3\linewidth]{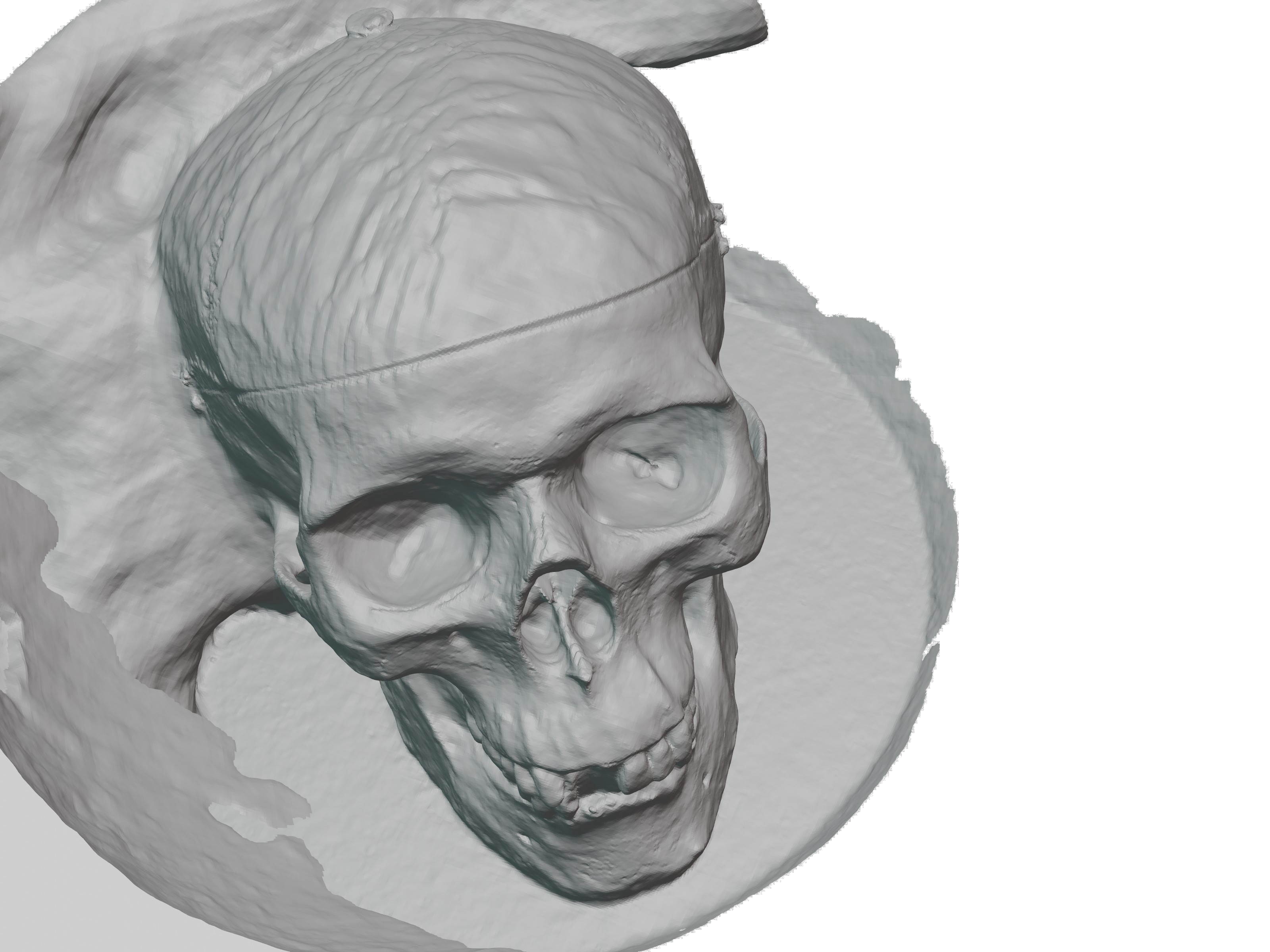} \\

        \rotatebox{90}{\parbox[t]{3cm}{\hspace*{\fill}Scan 118\hspace*{\fill}}}\hspace*{5pt} & 
        \includegraphics[trim=2cm 2cm 2cm 2cm, clip, width=0.3\linewidth]{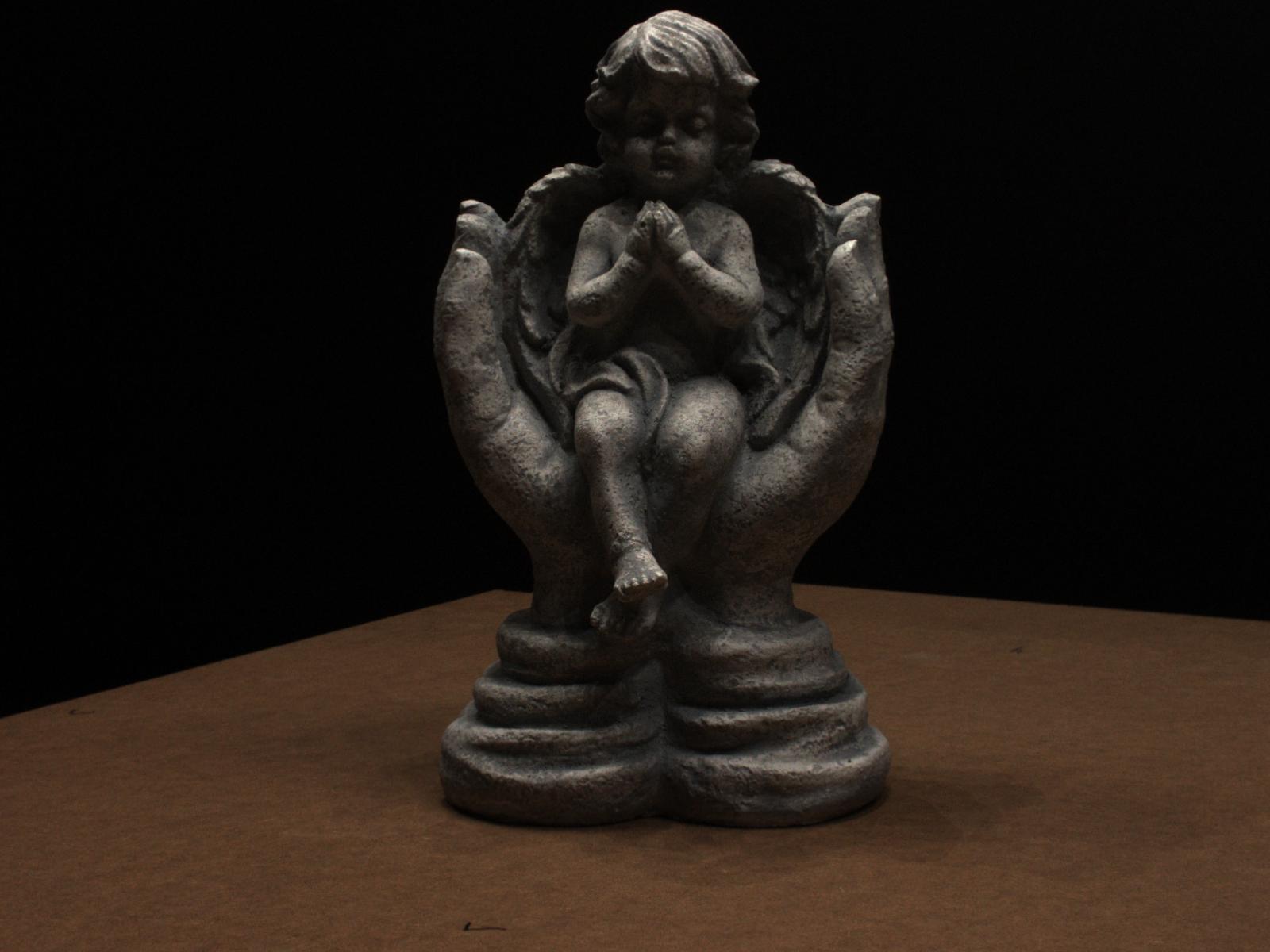} & 
        \includegraphics[trim=4cm 4cm 4cm 4cm, clip, width=0.3\linewidth]{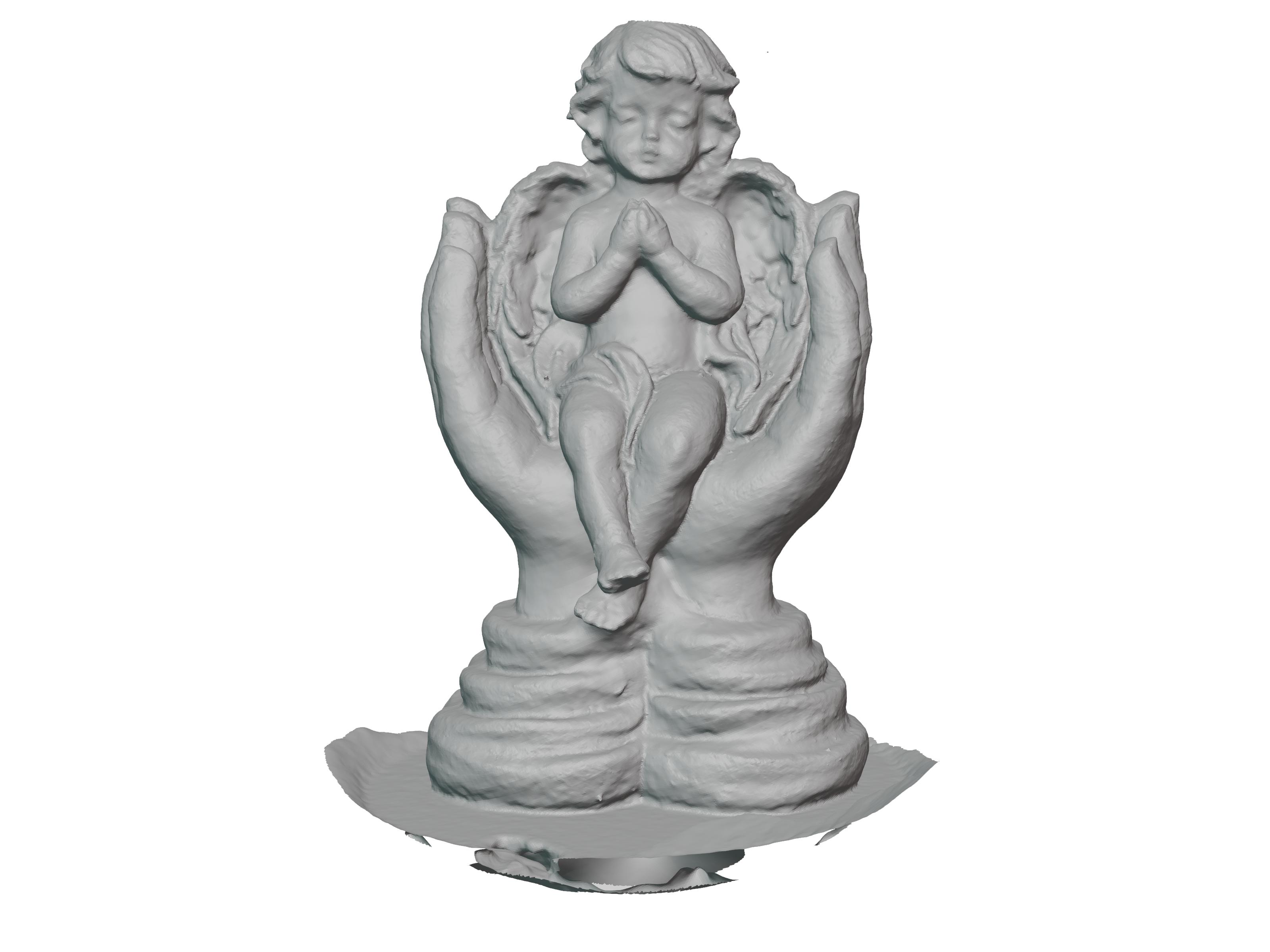} & 
        \includegraphics[trim=1cm 1cm 1cm 1cm, clip, width=0.30\linewidth]{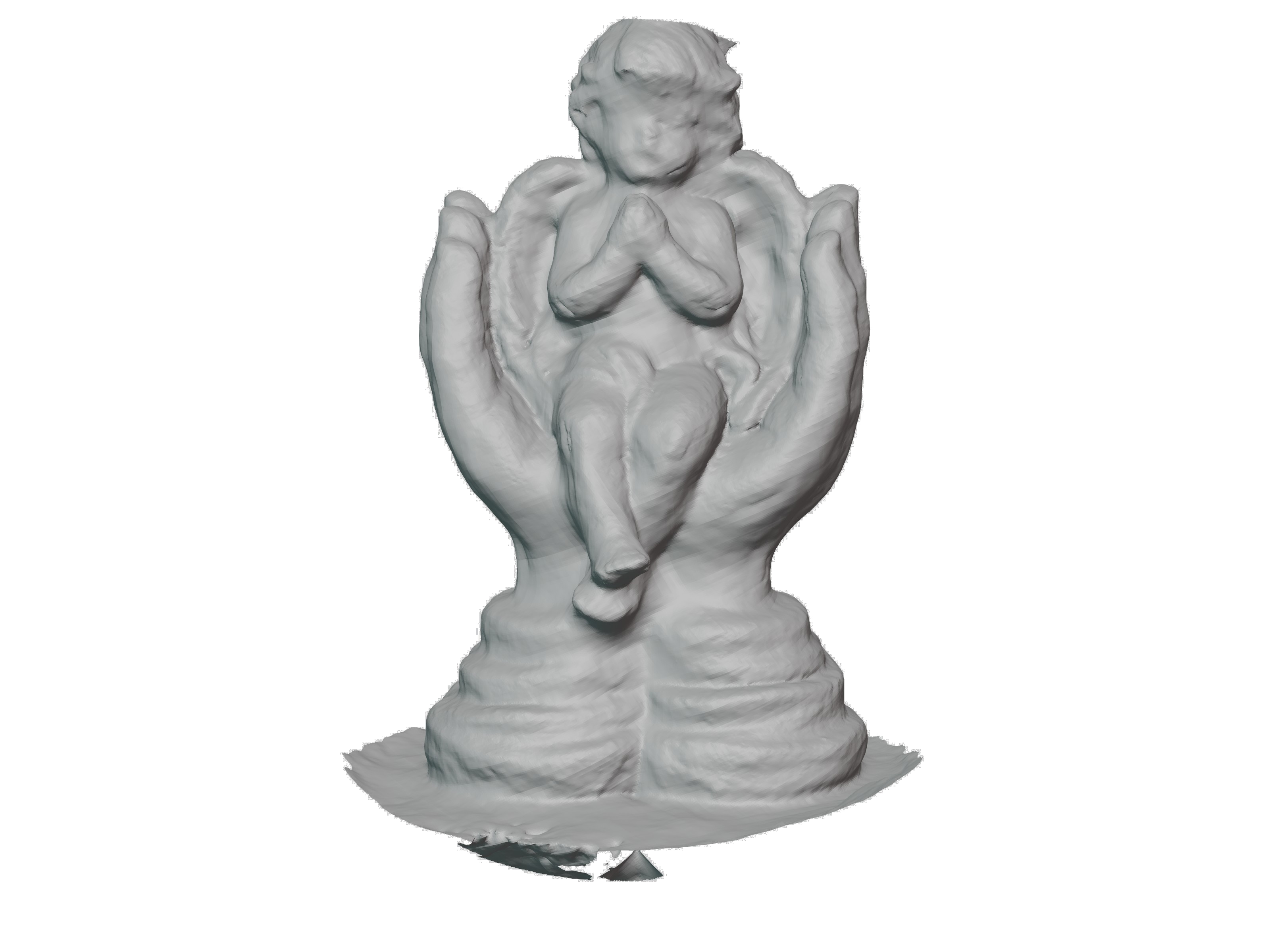} \\

    \end{tabular}
    \caption{Qualitative DTU results. Our approach improves surface details as well as surface accuracy of coarser components.}
    \label{fig:dtu_qual}
\end{figure*}

\section{Experiments}
\label{sec:exp}
We build our solution from the official Neuralangelo \cite{li2023neuralangelo} repository in Pytorch, and will release our code after acceptance. 
We evaluate our method in surface reconstruction tasks and report the performance in two standard benchmarks, DTU~\cite{aanaes2016large} and Tanks and Temples (TNT) \cite{Knapitsch2017}. 
All models are trained using a single Tesla V100 32GB GPU.

\paragraph{Network Specifications} For the SDF and RGB networks, we exactly follow the network specifications from Neuralangelo. Namely, we use a multi-layer perceptron (MLP) with one hidden layer of 256 units and softplus activation functions \cite{zheng2015improving}. For the positional encoding, we use a multi-resolution hash-grid of 16 levels, ranging from $32^{3}$ to $2048^{3}$, with a feature dimension of 8 and a hash dictionary size of $2^{22}$. For the RGB network, we use an MLP with 4 hidden layers of 256 units with ReLU activation functions \cite{nair2010rectified}. We encode view direction using 3 degrees of spherical harmonics. For the additional spatial mask network we use a small multi-layer perceptron with one hidden layer of 16 units with softplus activation functions. As a positional encoding, we use a small hash-grid of 8 levels ranging from $32^{3}$ to $2048^{3}$ units, but with feature dimension of 4 and a hash dictionary size of $2^{18}$. As a result, our model only uses marginally more parameters than Neuralangelo (374M versus 366M).
\subsection{DTU Results}

\begin{table*}[h!]
\centering
\label{table:tnt}
\resizebox{0.75\linewidth}{!}{
\begin{tabular}{lccccccccc}
\toprule
Method & Barn & Caterpillar & Courthouse & Ignatius & Meetingroom & Truck & Mean \\
\midrule
COLMAP \cite{schoenberger2016sfm} & 0.22 & 0.18 & 0.08 & 0.02 & 0.08 & 0.35 & 0.15 \\
NeuS \cite{wang2023neus} & 0.55 & 0.01 & 0.11 & 0.22 & 0.19 & 0.19 & 0.21 \\
Neuralangelo \cite{li2023neuralangelo} & 0.70 & 0.36 & 0.28 & \textbf{0.89} & \textbf{0.32} & 0.48 & 0.50 \\
\midrule
Neuralangelo \textdagger \cite{li2023neuralangelo} & 0.65 & 0.37 & 0.21 & 0.86 & 0.27 & 0.47 & 0.47 \\
Ours & \textbf{0.74} & \textbf{0.38} & \textbf{0.30} & \textbf{0.89} & 0.31 & \textbf{0.53} & \textbf{0.53} \\
\bottomrule
\end{tabular}
} 
\caption{Quantitative results on TNT dataset. We report F1 scores as attain the superior reconstructions. We include both the metrics of Neuralangelo reported in the paper, as well as the results obtained using the official repository indicated with a \textdagger.}
\end{table*}

\begin{figure*}[h!]
    \centering
    \begin{tabular}{cccc}
        & Input & Ours & Neuralangelo   \\ 
        
        \rotatebox{90}{\parbox[t]{3cm}{\hspace*{\fill} Truck \hspace*{\fill}}}\hspace*{5pt} & 
        \includegraphics[width=0.3\linewidth]{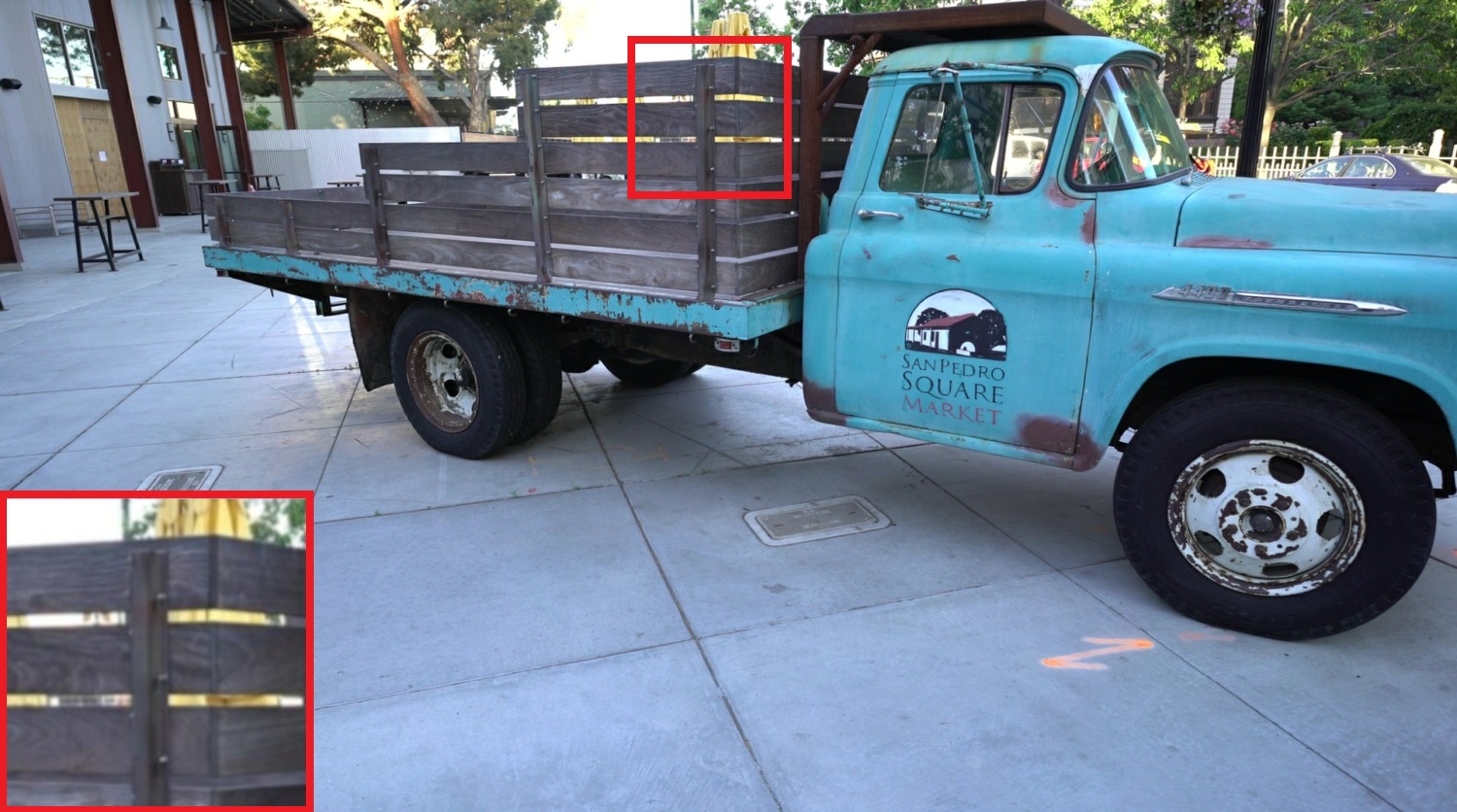} & 
        \includegraphics[width=0.3\linewidth]{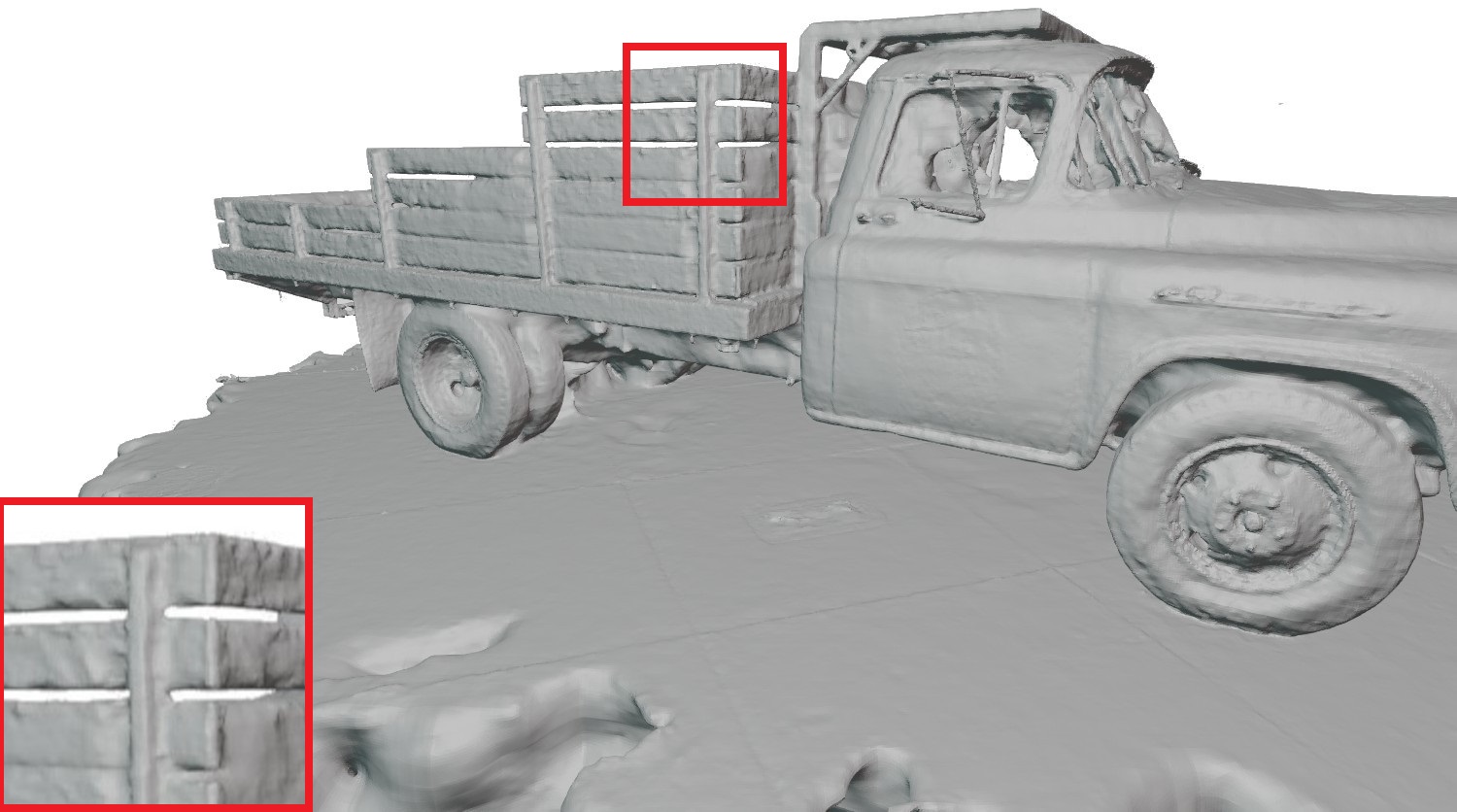} & 
        \includegraphics[width=0.3\linewidth]{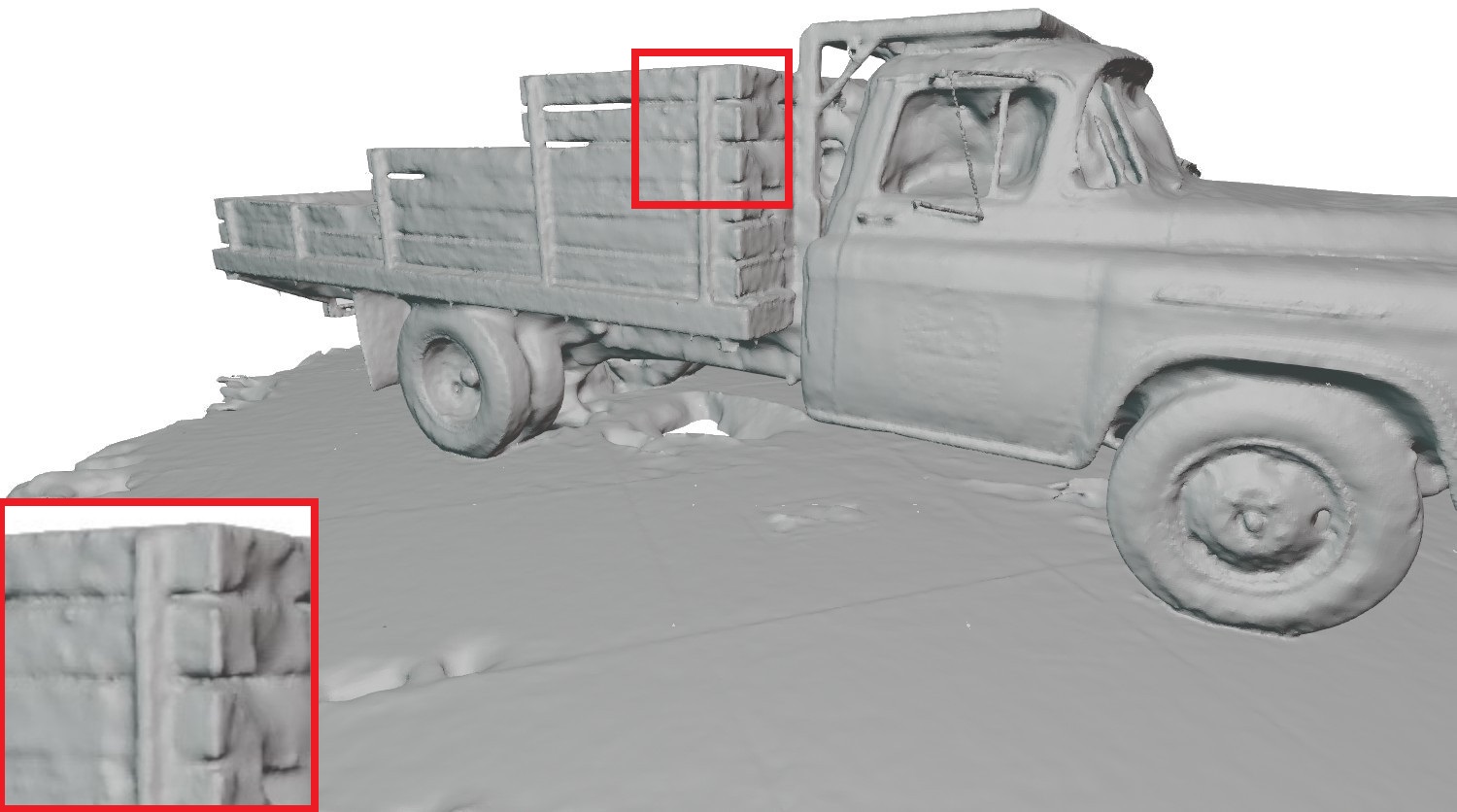} \\
        
        \rotatebox{90}{\parbox[t]{3cm}{\hspace*{\fill} Courthouse \hspace*{\fill}}}\hspace*{5pt}& 
        \includegraphics[width=0.3\linewidth]{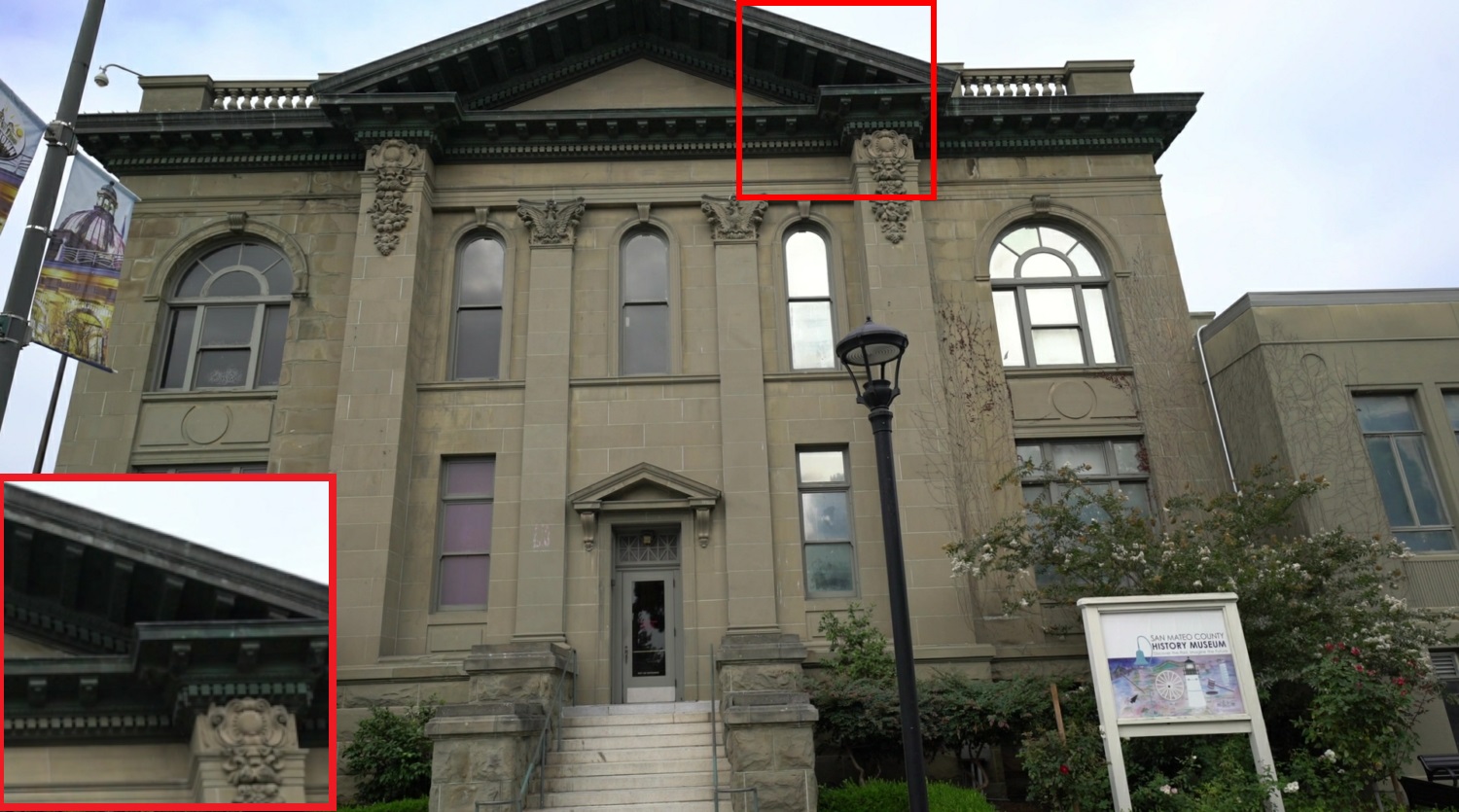} & 
        \includegraphics[width=0.3\linewidth]{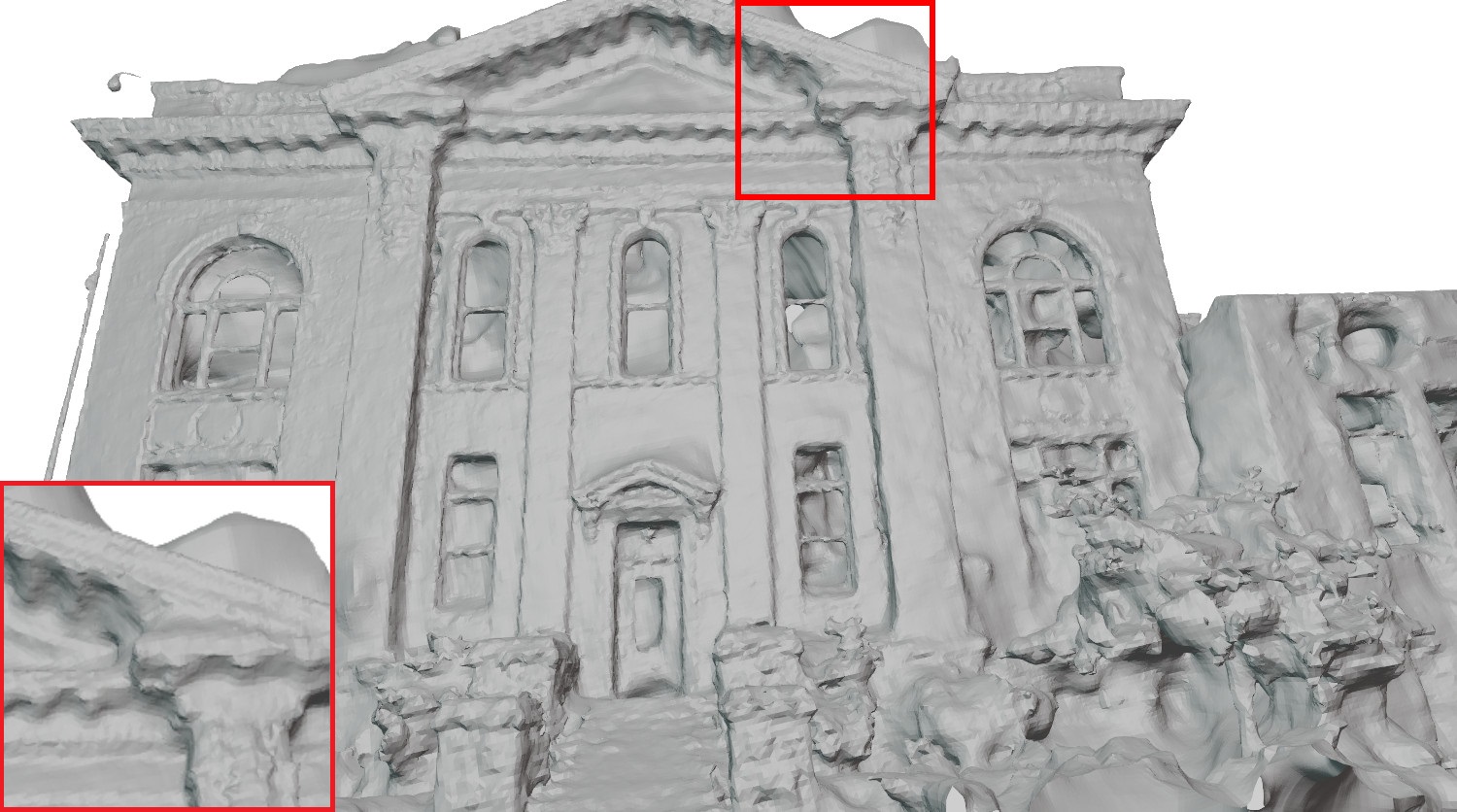} & 
        \includegraphics[width=0.3\linewidth]{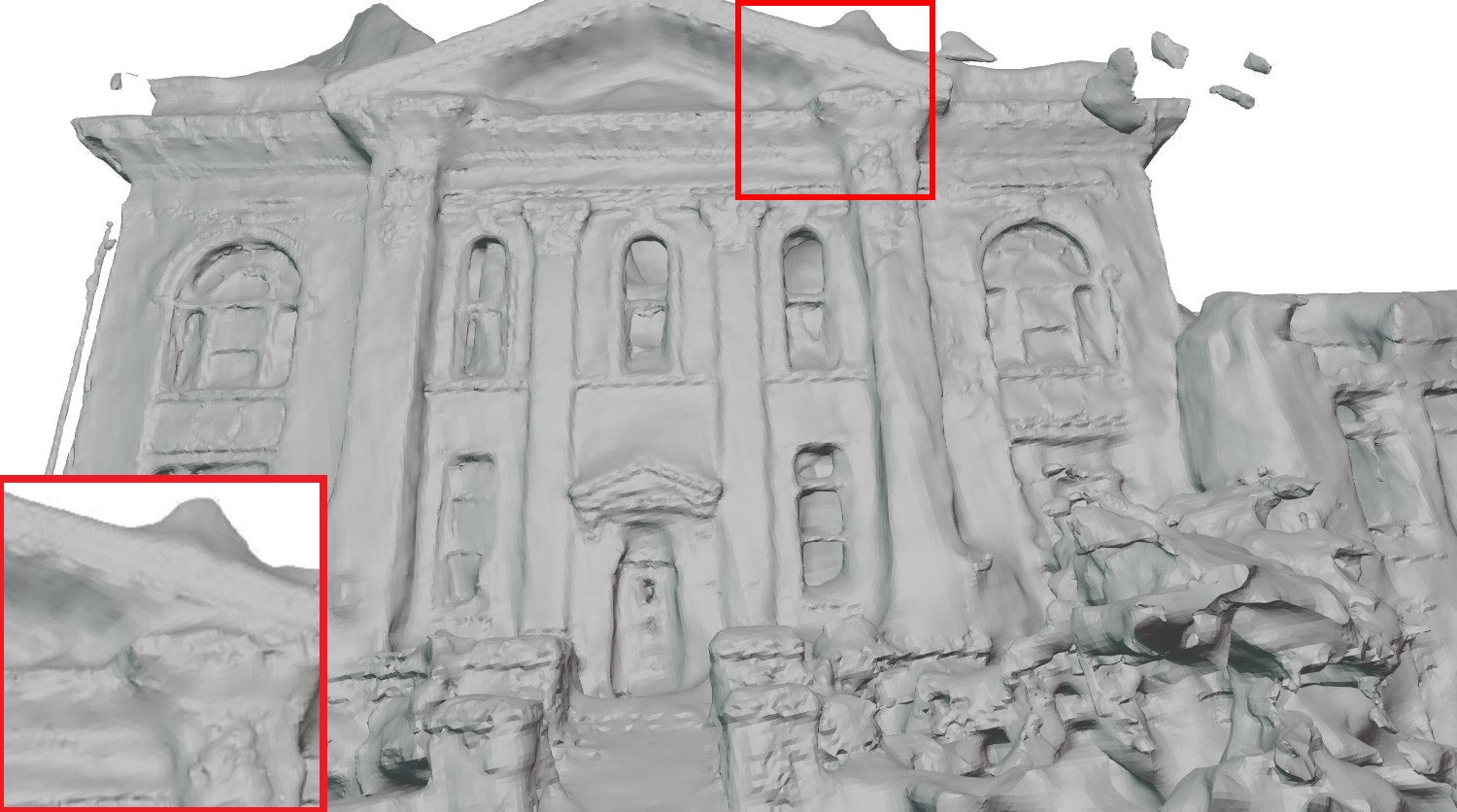}\\
        
        \rotatebox{90}{\parbox[t]{3cm}{\hspace*{\fill} Barn  \hspace*{\fill}}}\hspace*{5pt} & 
        \includegraphics[width=0.3\linewidth]{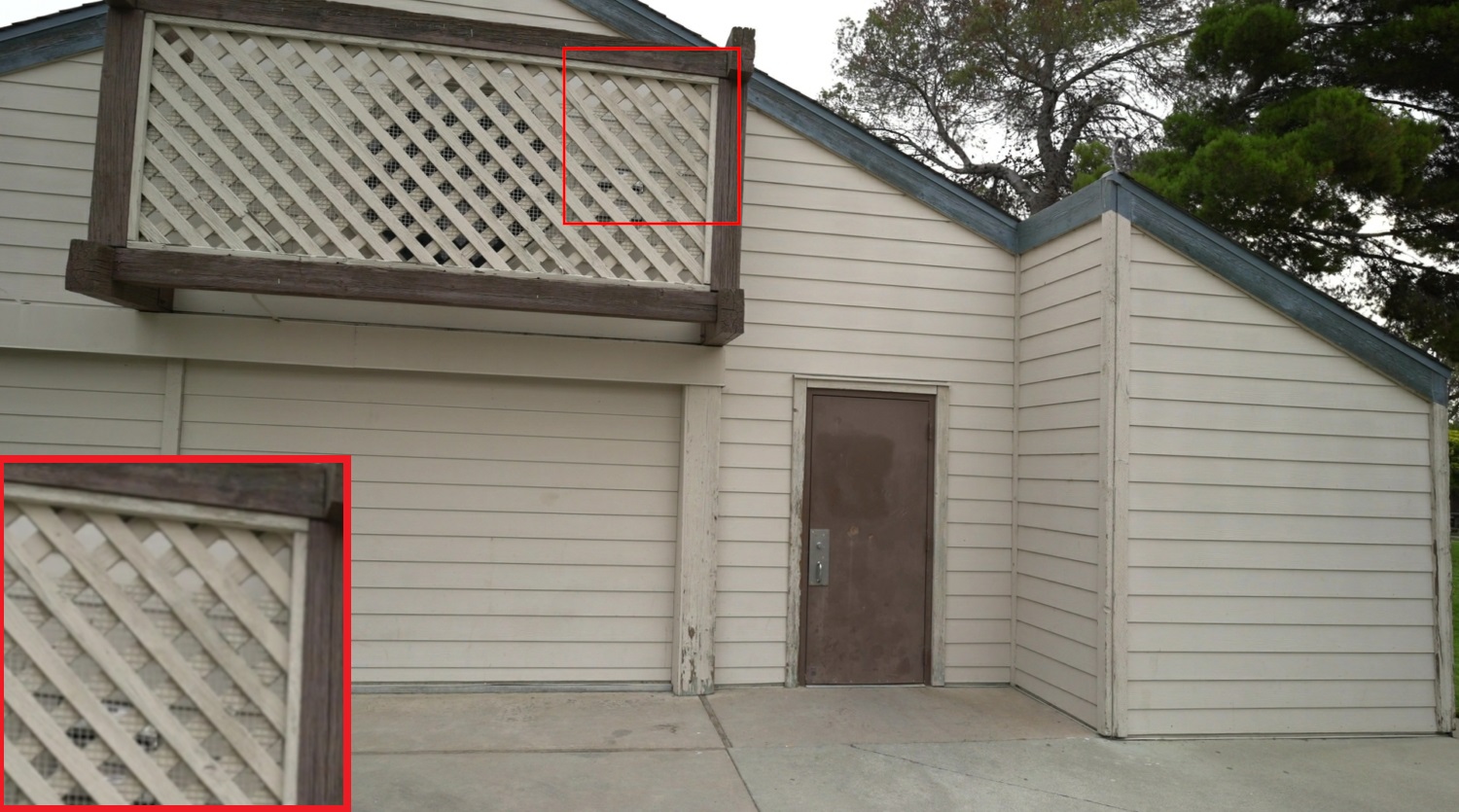} & 
        \includegraphics[width=0.3\linewidth]{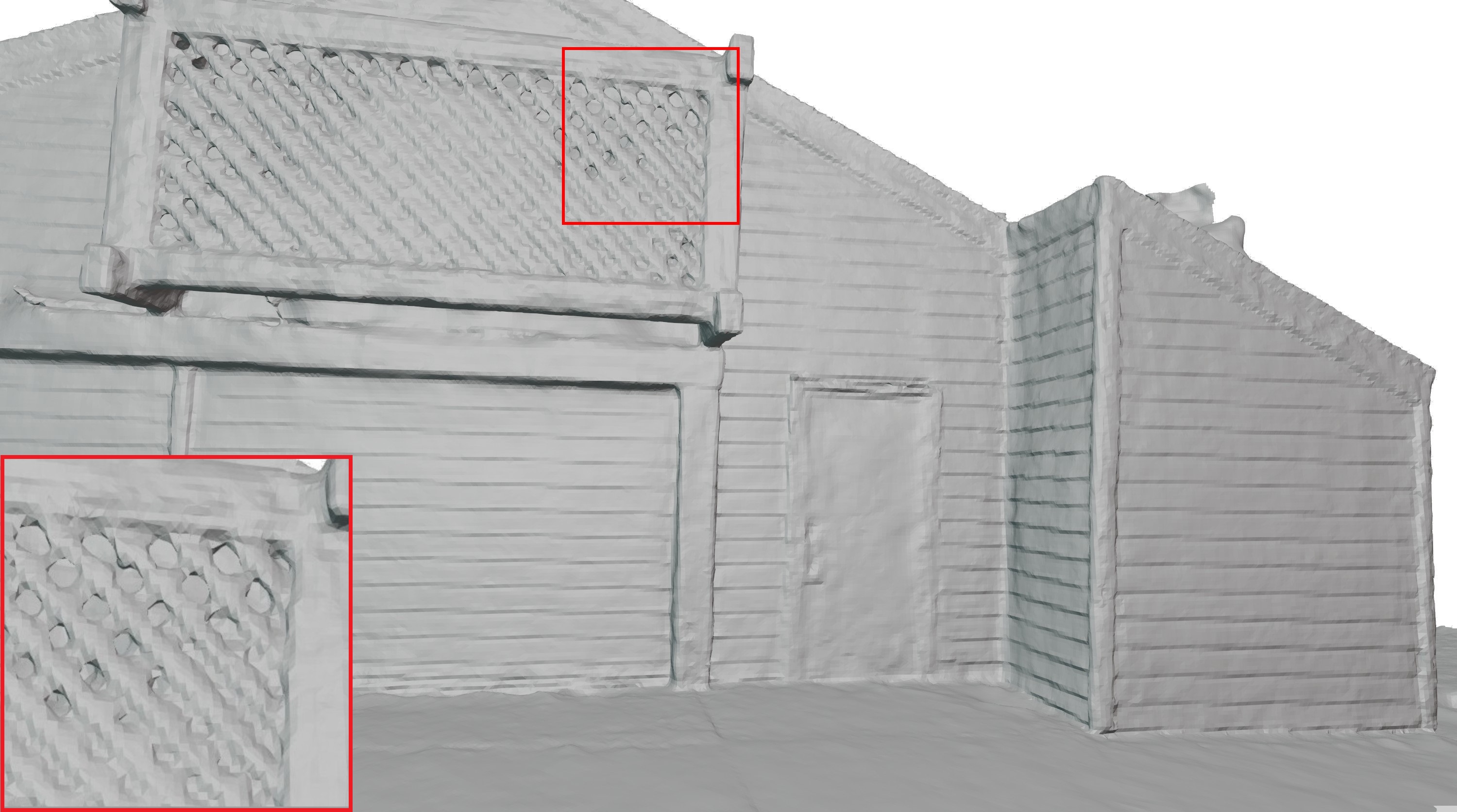} & 
        \includegraphics[width=0.3\linewidth]{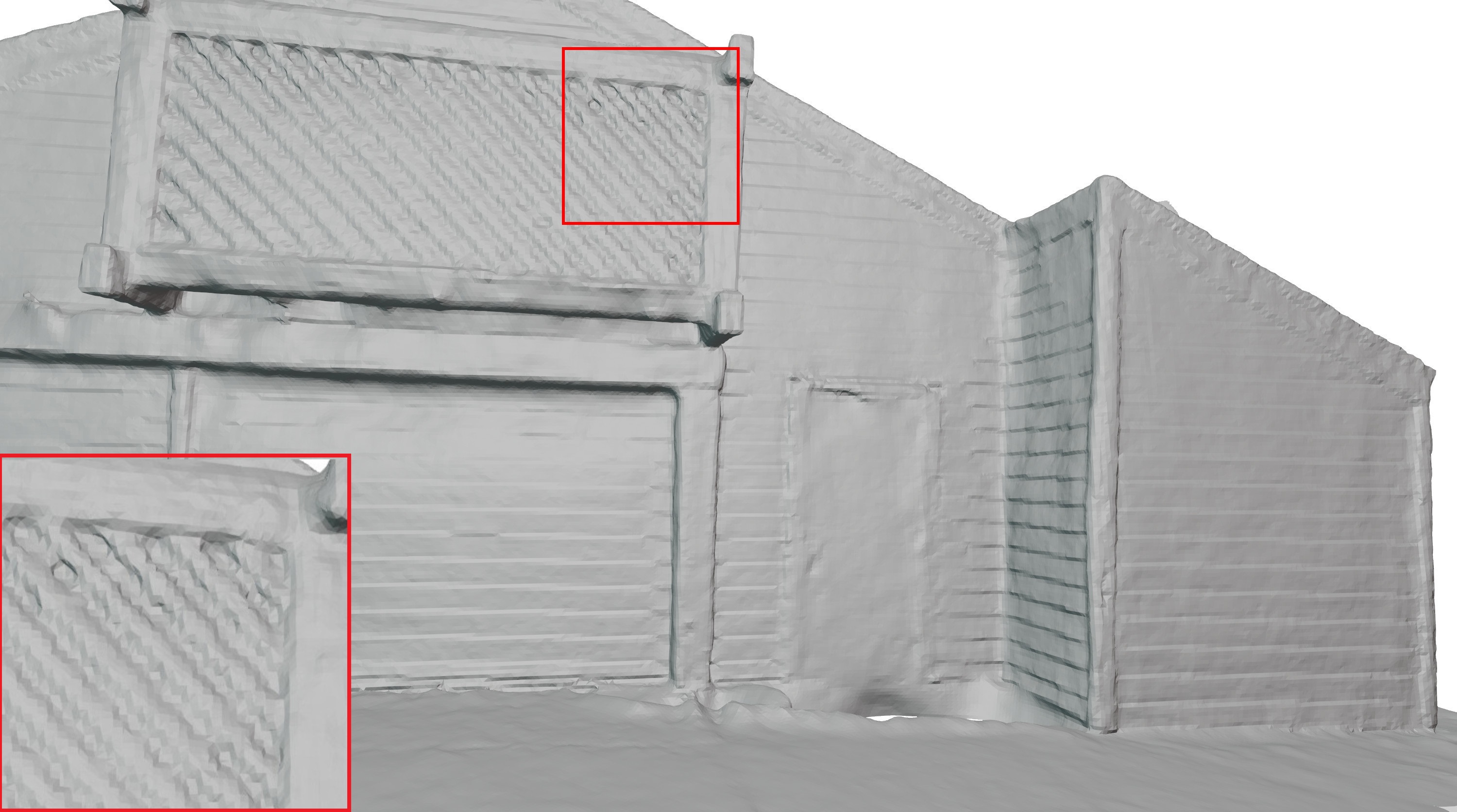} \\

    \end{tabular}
    \caption{Qualitative Tanks and Temples results. Our approach produces improved surface accuracy, attaining generally cleaner surfaces with improved fine details.}
    \label{fig:tnt_qual}
\end{figure*}

Following prior work, we use 15 scenes from the DTU dataset~\cite{aanaes2016large} for comparison. Each scene contains $1600 \times 1200$ resolution images, each paired with camera parameters and binary masks, from 49 or 64 poses. 
We compare the predicted surface reconstruction accuracy is to the provided ground truth point clouds that are obtained using a structured-light 3D scanner.
Overall, these scenes present difficulties for reconstruction algorithms due to the presence of non-Lambertian materials, intricate geometries, inconsistent lighting, and ambiguities in geometry and texture. We adhere to the official DTU evaluation protocol \cite{aanaes2016large} in to compute Chamfer-L1 scores against ground truth point clouds. The quantitative results are detailed in \Cref{tab:dtu}.\\

\begin{table*}[h!]
    \centering
    \resizebox{1\linewidth}{!}{
    \begin{tabular}{lcccccccccccccccc}
        \toprule
        Method & 24 & 37 & 40 & 55 & 63 & 65 & 69 & 83 & 97 & 105 & 106 & 110 & 114 & 118 & 122 & Mean \\
        \midrule
        NeRF \cite{mildenhall2021nerf}& 1.90 & 1.60 & 1.85 & 0.58 & 2.28 & 1.27 & 1.47 & 1.67 & 2.05 & 1.07 & 0.88 & 2.53 & 1.06 & 1.15 & 0.96 & 1.49 \\
        VolSDF \cite{yariv2021volume} & 1.14 & 1.26 & 0.81 & 0.49 & 1.25 & 0.70 & 0.72 & 1.29 & 1.18 & 0.70 & 0.66 & 1.08 & 0.42 & 0.61 & 0.55 & 0.86 \\
        NeuS \cite{wang2023neus}& 1.00 & 1.37 & 0.93 & 0.43 & 1.10 & 0.65 & 0.57 & 1.48 & 1.09 & 0.83 & 0.52 & 1.20 & 0.35 & 0.49 & 0.54 & 0.84 \\
        HF-NeuS\cite{wang2022hf} & 0.76 & 1.32 & 0.70 & 0.39 & 1.06 & 0.63 & 0.63 & 1.15 & 1.12 & 0.80 & 0.52 & 1.22 & 0.33 & 0.49 & 0.50 & 0.77 \\
        Neuralangelo \cite{li2023neuralangelo}  & \textbf{0.37} & 0.72 & \textbf{0.35} & \textbf{0.35} & 0.87 & 0.54 & \textbf{0.53} & 1.29 & 0.97 & 0.73 & 0.47 & 0.74 & \textbf{0.32} & 0.41 & \textbf{0.43} & 0.61 \\
        \midrule
        Neuralangelo \textdagger \cite{li2023neuralangelo} & 0.42 & 0.92 & 0.40 & 0.41 & 1.41 & 0.95 & 1.89 & 1.42 & 1.80 & 0.80 & 0.44 & 1.13 & 0.35 & 1.02 & 1.07 & 0.96 \\
        Ours & \textbf{0.37} & \textbf{0.65} & \textbf{0.35} & \textbf{0.35} & \textbf{0.79} & \textbf{0.50} & 0.54 & \textbf{1.20} & \textbf{0.72} & \textbf{0.70} & \textbf{0.39} & \textbf{0.72} & \textbf{0.32} & \textbf{0.40} & 0.44 & \textbf{0.56} \\
        \bottomrule
    \end{tabular}
    }
    \caption{Quantitative results on DTU dataset. We report chamfer distance (in millimeters). Our spatially-adaptive encodings achieve the best surface reconstruction accuracy. We include both the metrics of Neuralangelo reported in the paper, as well as the results obtained using the official repository indicated with a \textdagger.}
    \label{tab:dtu}
\end{table*}

\begin{figure*}[h!]
    \centering
    \begin{tabular}{cccc}
        & Low-Frequency & Mid-Frequency & High-Frequency \\

        \rotatebox{90}{\parbox[t]{3cm}{\hspace*{\fill}DTU Scan 37 \hspace*{\fill}}}\hspace*{5pt} & 
        \includegraphics[width=0.3\linewidth]{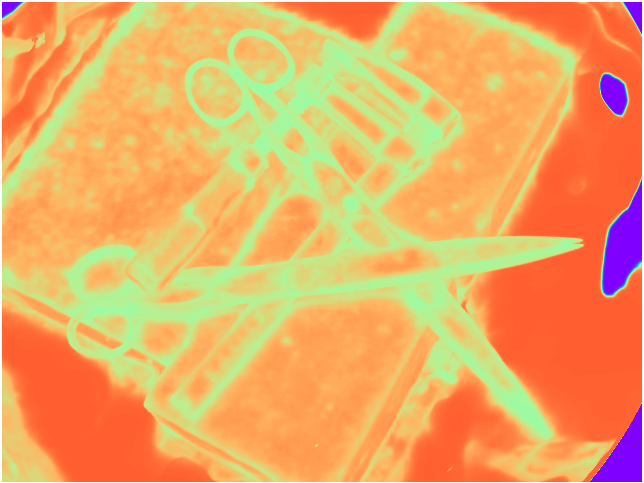} & 
        \includegraphics[width=0.3\linewidth]{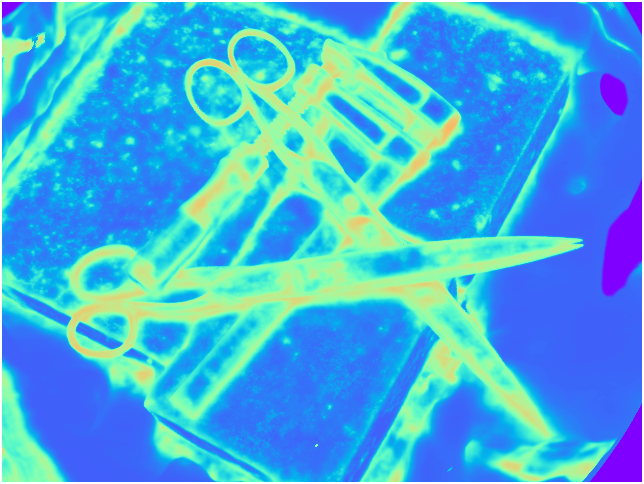} & 
        \includegraphics[width=0.3\linewidth]{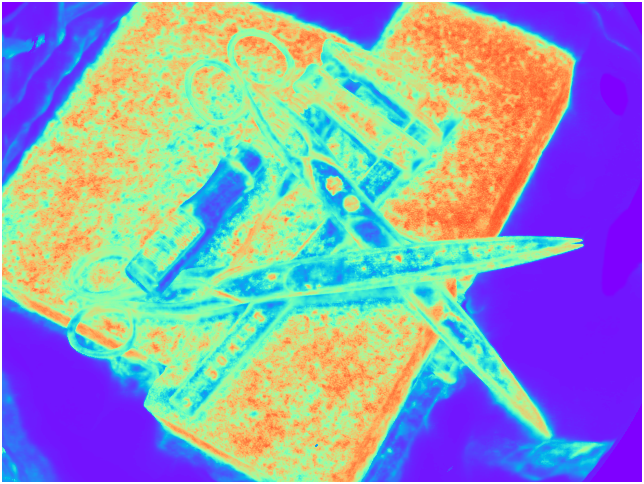} \\ 
        
        \rotatebox{90}{\parbox[t]{3cm}{\hspace*{\fill} DTU Scan 97 \hspace*{\fill}}}\hspace*{5pt}& 
        \includegraphics[width=0.3\linewidth]{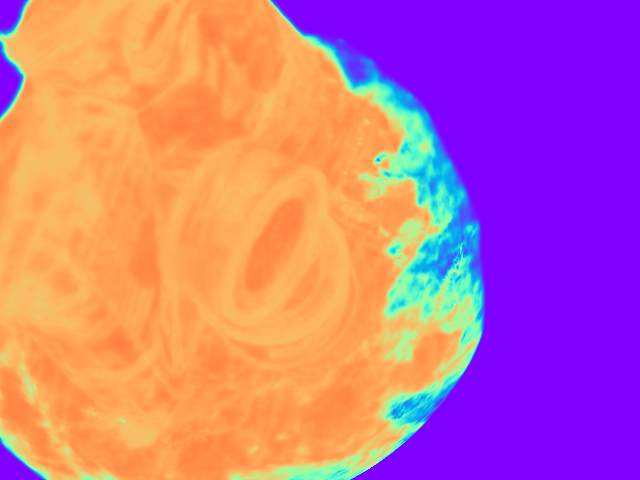} & 
        \includegraphics[width=0.3\linewidth]{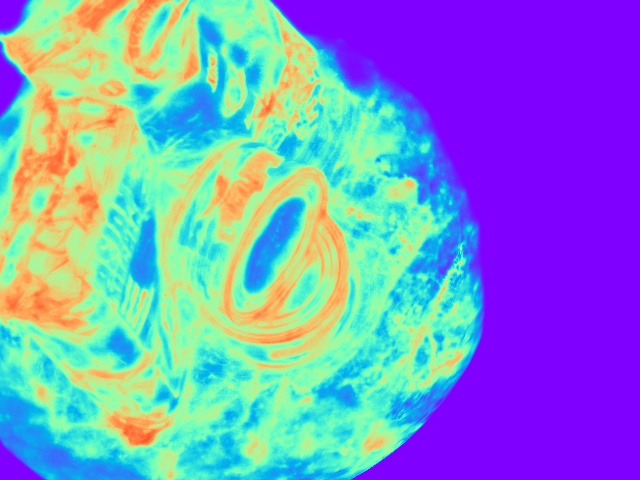} & 
        \includegraphics[width=0.3\linewidth]{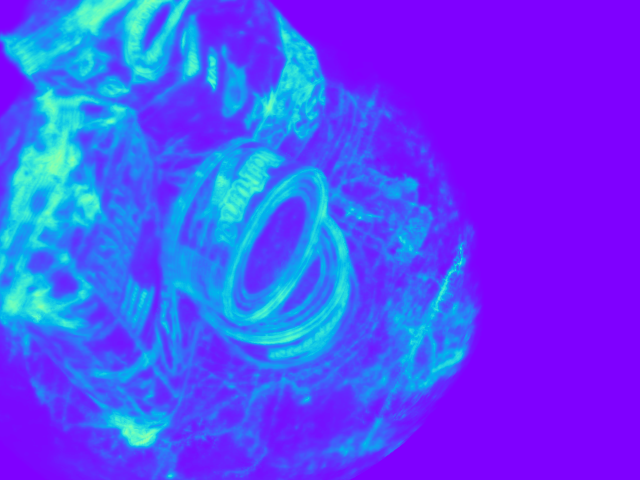} \\ 
        
        \rotatebox{90}{\parbox[t]{3cm}{\hspace*{\fill} TNT Truck \hspace*{\fill}}}\hspace*{5pt} & 
        \includegraphics[width=0.3\linewidth]{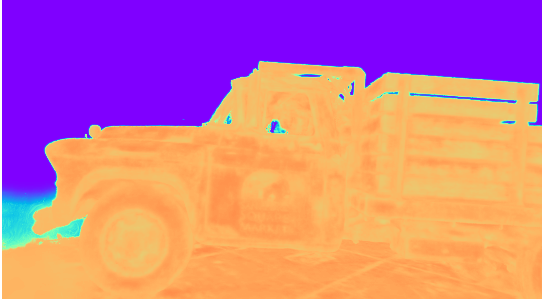} & 
        \includegraphics[width=0.3\linewidth]{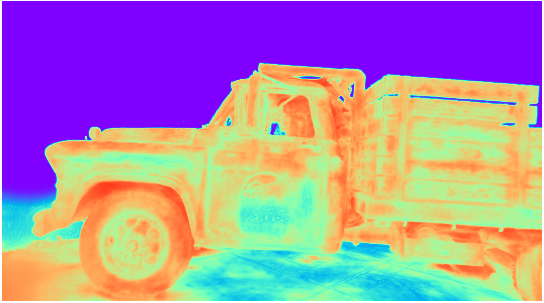} & 
        \includegraphics[width=0.3\linewidth]{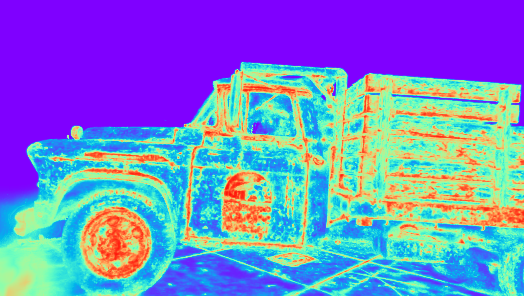} \\ 

        \rotatebox{90}{\parbox[t]{3cm}{\hspace*{\fill} TNT Barn\hspace*{\fill}}}\hspace*{5pt} & 
        \includegraphics[width=0.3\linewidth]{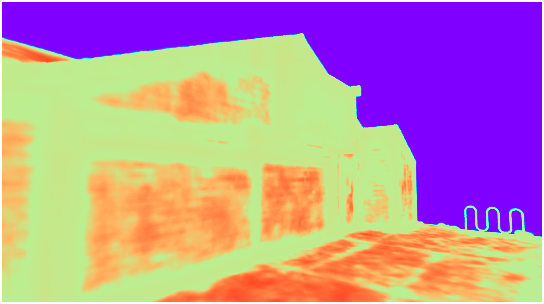} & 
        \includegraphics[width=0.3\linewidth]{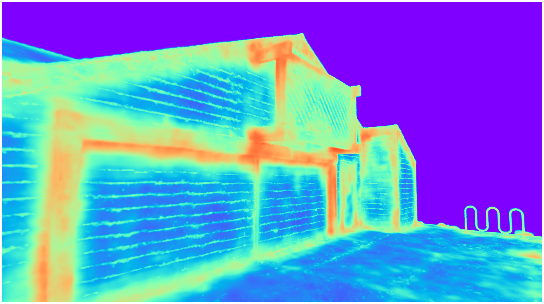} & 
        \includegraphics[width=0.3\linewidth]{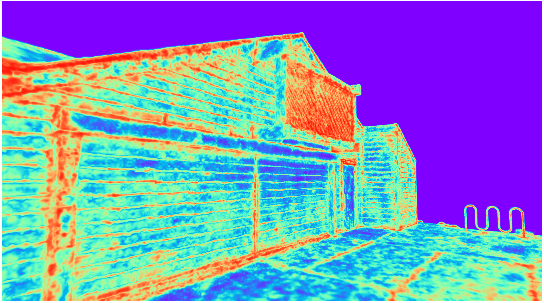} \\

         & 
         & 
        \includegraphics[width=0.3\linewidth]{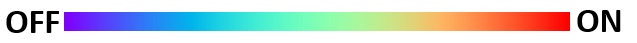} & 
           \\

    \end{tabular}
    \caption{Renderings of spatial mask heat maps. Red indicates the corresponding spatial masks take the value 1. Blue indicates the mask values are 0 and hence features stored on the corresponding grid resolutions are ignored in the final encoding.}
    \label{fig:masks}
\end{figure*}

\subsection{Tanks and Temples Results}
We further report on the indoor/outdoor image-based 3D reconstruction dataset Tanks and Temples \cite{Knapitsch2017}. Ground truth point clouds were obtained using an industrial laser scanner. We use the offical Tanks and Temples (TNT) evaluation script to compute F1 accuracy metrics. The results are detailed in \Cref{table:tnt}\\

\textbf{Discussion} Our spatially-adaptive hash-grids achieve superior quantitative performance on both DTU and as well as Tanks and Temples datasets. Qualitatively, we see less noise in low-frequency regions than Neuralangelo, for example the top of the skull or the blades of the scissors (see \Cref{fig:dtu_qual}). We also see improved surface details (see \Cref{fig:tnt_qual}) across TNT scenes.\\

In \Cref{fig:masks} we present heat map renderings of the learned spatial masks $s(\textbf{x})$ for each scene. Images were generated by volume rendering the spatial masks. We consider the first 8 components of the spatial mask the ``Low-Frequency" regime, as they correspond to the 8 coarsest grids which capture lower frequency geometry. The maximum is computed over these components, and then volume rendered into images using the density field induced by the final learned SDF. Similarly, we consider ``Mid-Frequency" and ``High-Frequency" to be components 8-14 and 15-16 respectively. Interestingly, for the high-frequency heat maps we observe segmentation of fine details such as edges or gratings which necessitate the finest highest resolution of hash-grid to be resolved. For example, in the scissors scene (DTU Scan 37) we observe the highest resolution grids being used most prominently on the details of scissors as well as the very rough ceramic block (see \Cref{fig:dtu_qual} for groundtruth). Meanwhile for the smooth handle of the tool, a high-frequency embedding is not used and the hence the network has a bias towards low-frequency geometry in this region. This could explain our methods ability to correctly reconstruct this region of the scene despite specularities, whilst Neuralangelo fails and results in high-frequency concavities. \\

On the TNT scenes, we observe improvements over Neuralangelo in reconstructing fine details precisely in regions where the spatial mask network has activated the highest resolution grids (whilst partially silencing the contribution of mid and low-frequency encodings). For example, the grating of the Barn, wheels of the truck (see \Cref{fig:tnt_qual} for zooms of these areas). Note that no regularization is used to encourage sparsity in the spatial mask field. Instead, our combination of progressive unveiling of high-resolution grids with our spatial varying mask provides a mechanism to only use these embeddings where necessary and avoid unwanted noise from hash-collisions.

\subsection{Ablation}
In this section we present ablations of the design of the spatial mask network. 

\begin{figure}[h!]
    \centering
    \begin{tabular}{cc}
        \includegraphics[width=0.45\linewidth]{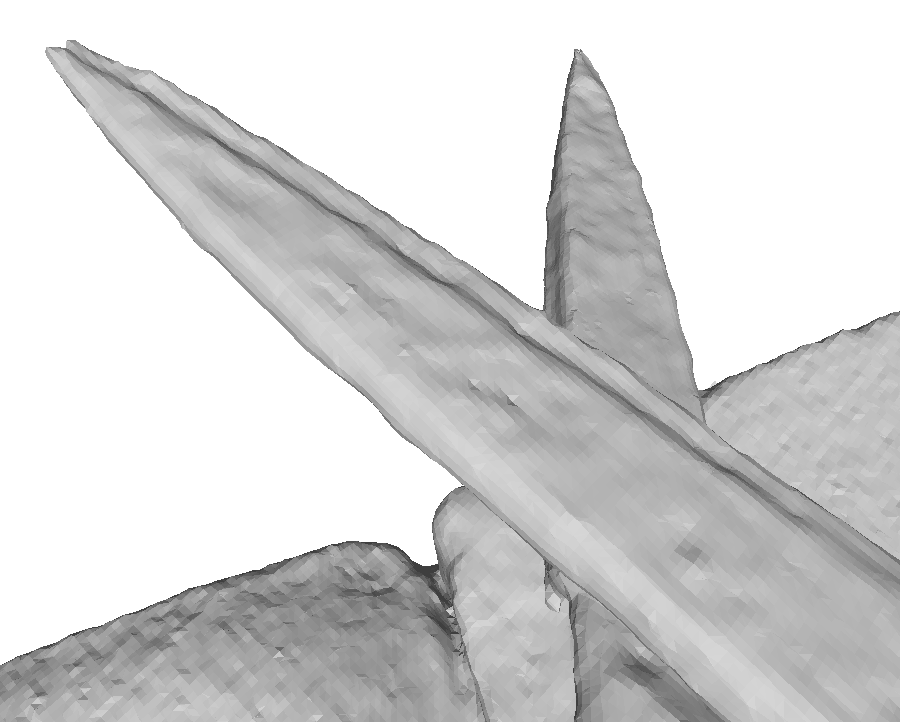} & 
        \includegraphics[width=0.45\linewidth]{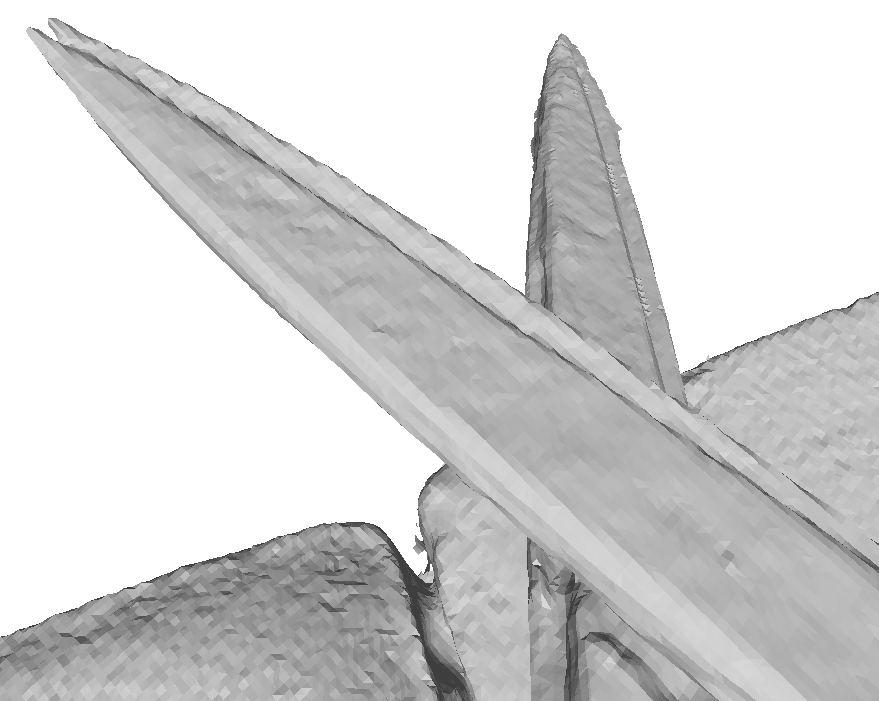} \\
        0-6 & 5-11 (Ours) \\ 
    \end{tabular}
    \caption{Close-up comparison of details in scene 37. Left: spatial mask network with coarse hash-grid. Right: spatial mask network with default $[d_{min}, d_{max}] = [5, 11]$ grid resolutions.  When using the coarsest mask hash-grid, the sharp features of the scissor blade are lost.}
    \label{fig:mask_grid_res}
\end{figure}

\paragraph{Spatial Mask Hash-Grid Resolution} In \Cref{table:ablation_mask_res} we test various resolutions of multi-resolution hash-grid. Minimum and maximum grid resolutions are selected as $2^{d_{min}}$ and $2^{d_{max}}$. At its coarsest level, we observe a decline in surface accuracy, both quantitatively and qualitatively (see \Cref{fig:mask_grid_res}). This highlights the importance of the spatial mask network in locally modulating features from different grids. 

\begin{table}[h!]
\centering
\label{table:ablation_mask_res}
\resizebox{0.9\linewidth}{!}{
\begin{tabular}{l|ccccccccc}
\toprule
$d_{min}$\,- \,$d_{max}$ & 0-6 & 2-8 & 4-10 & 5-11\\
\midrule
 Chamfer & 0.670 & 0.655 & \textbf{0.648} & 0.654 \\
\bottomrule
\end{tabular}
} 
\caption{Quantitative results of spatial mask hash-grid resolution ablation. We report Chamfer distance (mm).}
\end{table}

\begin{figure}[h!]
    \centering
    \begin{tabular}{ccc}
        \includegraphics[width=0.32\linewidth]{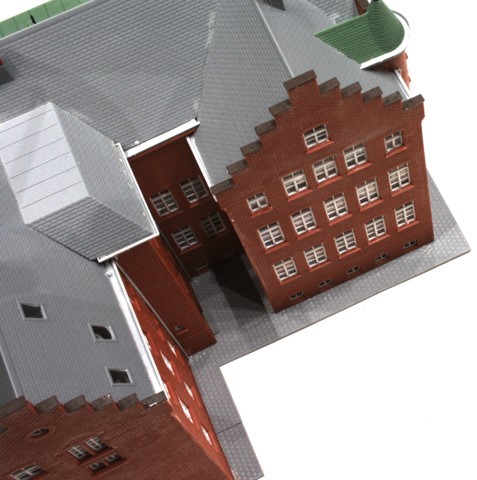} & 
        \includegraphics[width=0.32\linewidth]{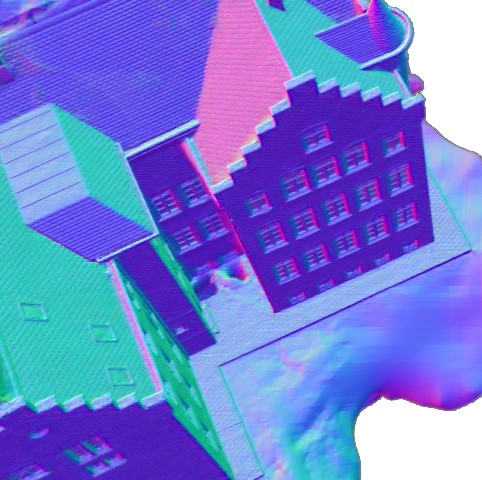} & 
        \includegraphics[width=0.32\linewidth]{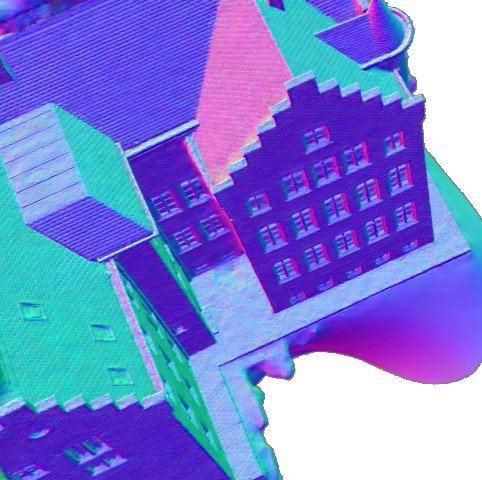} \\
        Input & Without $ w_{\text{curv}}$ & With $ w_{\text{curv}}$ \\ 
    \end{tabular}
    \caption{Ablation results of models with and without curvature regularization. Without this prior, we see unwanted artifacts emerge on the roof and concavity of scene 24.}
    \label{curv_abl}
\end{figure}

\vspace{-8mm}
\paragraph{Curvature Regularization}
In \Cref{curv_abl} we ablate the use of curvature regularization. Similarly to Neuralangelo, we find curvature regularization is a useful prior to preserve surface continuity and encourage smooth, aesthetic geometry. 

\begin{figure}[h!]
    \centering
    \begin{tabular}{cc}
        \includegraphics[width=0.45\linewidth]{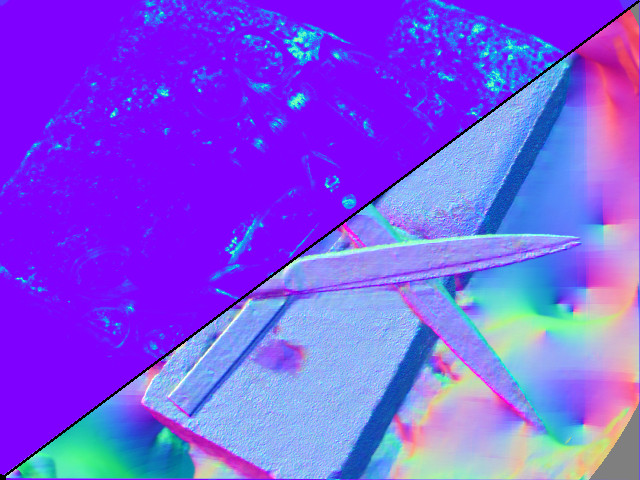} & 
        \includegraphics[width=0.45\linewidth]{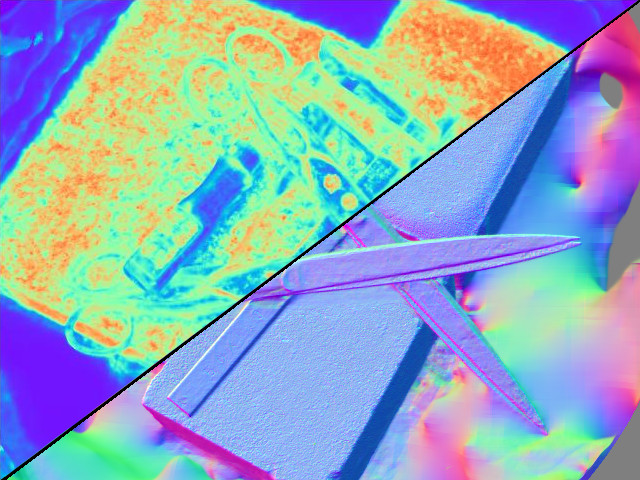} \\
        Softmax, CD: 0.74 & Sigmoid, \textbf{CD: 0.65} \\ 
    \end{tabular}
    \caption{Ablation results of models using sigmoid and softmax activations. The top left half shows the high frequency mask and the bottom right shows the geometry normal map.}
    \label{fig:softmax}
\end{figure}

\paragraph{Softmax Activation} Another possible choice of activation fuction for the masks is the softmax, forcing the network to (softly) select a specific resolution of hash grid. In \Cref{fig:softmax} we compare masks and reconstructions of models using sigmoid and softmax activation functions. The use of softmax results in the model opting to select the lowest frequency masks and hence features from the finest grids being mostly ignored, degrading surface reconstruction performance.

\section{Limitations and Conclusions}
\label{sec:conc}
Various positional encoding strategies remain in the literature, however, issues with capturing fine surface details and achieving stable optimization persist. 
Our proposed spatially-adaptive formulation of multi-resolution hash grids addresses these challenges, and achieves state-of-the-art performance across multiple datasets. Despite this, we inherit the same limitations of Neuralangelo and other grid-based methods. Namely, the need to explicitly store features of the hash-grids requires a lot of memory ($\approx 4.2$ GB). To address this, future work could consider using the masks to prune features from unused grids, thereby extending Hollow-NeRF~\cite{xie2023hollownerfpruninghashgridbasednerfs} to allow a more nuanced per-grid pruning strategy. Our method also shares a limitation of hash-grid-based encodings noted by Neuralangelo \cite{li2023neuralangelo}. Specifically, they tend to perform less favorably in highly reflective scenes compared to frequency-based encodings. A promising direction for future work could be to integrate our approach with spatially-adaptive view direction encodings \cite{verbin2022ref, Han_2023_CVPR} which are designed to handle spatially varying reflective properties. This fully spatially-adaptive model could be promising for scenes with both specular and Lambertian materials.

\clearpage
{\small
\bibliographystyle{ieee_fullname}
\bibliography{egbib}
}

\end{document}